\documentclass{article}

\PassOptionsToPackage{numbers}{natbib}
\usepackage[preprint]{neurips_2026}

\usepackage[utf8]{inputenc} 
\usepackage[T1]{fontenc}    
\usepackage{hyperref}       
\usepackage{url}            
\usepackage{booktabs}       
\usepackage{amssymb}
\usepackage{amsfonts}       
\usepackage{nicefrac}       
\usepackage{microtype}      
\usepackage{xcolor}         
\usepackage{graphicx}
\usepackage{amsmath}
\usepackage{makecell}
\usepackage{bbm}
\usepackage{paralist}
\usepackage{multirow}
\usepackage{subcaption}
\usepackage{graphicx}
\usepackage{todonotes}
\usepackage{fontawesome5}

\usepackage{tabularx}
\usepackage{array}

\title{Temporally-Resolved Token Attribution Reveals the Generation Dynamics of Diffusion Language Models}
\workshoptitle{DiffuLM Workshop at NeurIPS 2026}

\author{
  \textbf{Darpan Aswal$^{1}$} \& \textbf{C\'eline Hudelot$^{2}$} \\
  $^1$Universit\'e Grenoble Alpes, $^2$MICS, CentraleSupélec
  \\
  \texttt{darpanaswal@gmail.com}\\
  \makebox[\linewidth][c]{%
    \small
    \begin{tabular}{@{}c@{\hspace{3em}}c@{}}
      Code: \faGithub ~ \href{https://github.com/darpanaswal/DLIG}{darpanaswal/DLIG} &
      Models: {%
        \raisebox{-0.3em}{\includegraphics[height=1.3em]{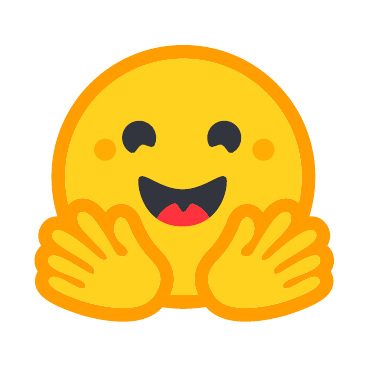}} \href{https://hf.co/collections/darpanaswal/dlig}{darpanaswal/DLIG}}
    \end{tabular}%
    }
}

\begin{document}
\maketitle
\begin{abstract}
This work presents Diffusion Layer Integrated Gradients (DLIG), a token attribution method for diffusion language models (DLMs) that extends Integrated Gradients (IG~\cite{sundararajan2017axiomatic}) to arbitrary layers and denoising steps. DLIG attributes a DLM's progressive commitment to a self-generated or fixed completion for an input prompt. We establish direct correspondences between DLIG and the IG axioms of completeness, implementation invariance, linearity, and symmetry preservation. As a lightweight complement to interventional analysis, DLIG provides an inexpensive first check of mechanistic hypotheses across the denoising trajectory. We demonstrate this on word-sense disambiguation, multi-hop graph reasoning, and sentence infilling, revealing how DLMs draw on inputs across positions, layers, and denoising steps.
\end{abstract}
\section{Introduction}
\label{sec:introduction}

Diffusion language models (DLMs) generate by iterative refinement rather than left-to-right prediction, unmasking a sequence over a trajectory of denoising steps~\cite{gong2025scaling, ye2508dream}. This changes what it means to explain a generation: the same output token is arrived at gradually and the computation supporting it is distributed over both network depth and denoising time. Explaining a DLM's output therefore requires locating support not only \emph{where} in the network it lives but \emph{when} along the trajectory it forms.

Yet post-hoc attribution for DLMs remains largely unaddressed. A recent taxonomy formalizes what an attribution contract for iterative generation should provide~\cite{nguyen2026attribution}, but no concrete method instantiates one: existing token-attribution tools carry no notion of a denoising step, while DLM interpretability has primarily adapted expensive interventional analyses such as causal tracing~\cite{yao2026timerome}.

We close this gap with \textbf{Diffusion Layer Integrated Gradients (DLIG)}. DLIG scores the model's commitment to its self-generated completion at a chosen layer and denoising step, and attributes that score to the prompt by integrating gradients along a path from a prompt-free baseline to the prompt-present state. Resolved over a grid of layers and steps, it yields a per-token attribution field exposing how output-driving support evolves through denoising. The construction admits an exact correspondence to Integrated Gradients~\cite{sundararajan2017axiomatic} applied at a fixed $(\text{layer},\text{step})$, which lets us carry the IG axioms over to DLIG. Crucially, this correspondence holds at a \emph{single} coordinate; DLIG's value comes from assembling these axiom-preserving scores across the layer--step grid to expose trajectory structure that single-pass attribution cannot reveal. This temporal resolution parallels recent causal-tracing approaches for DLMs~\cite{yao2026timerome}, while providing a relatively light-weight first-pass analysis.

We demonstrate DLIG on word-sense disambiguation, multi-hop reasoning, and sentence infilling with DiffuGPT-M~\cite{gong2025scaling}, using it to test concrete hypotheses about \emph{where} and \emph{when} output-driving evidence emerges. \S~\ref{sec:dlig} defines DLIG and extends the IG axioms to it (\S~\ref{sec:axioms}); \S~\ref{sec:wic}, \S~\ref{sec:prosqa}, and \S~\ref{sec:rocstories} report the three tasks, with implementation checks and full statistics in Appendices~\ref{app:impl-checks} and~\ref{app:stats}. This positions DLIG as a lightweight first check that can identify or rule out candidate mechanisms before more expensive causal interventional analyses.
\section{Background and Related Work}
\label{sec:background}
\paragraph{Diffusion Language Models (DLMs).} 
Diffusion, originating in nonequilibrium thermodynamics~\cite{sohl2015deep}, has become the central algorithm behind modern image generation models~\cite{ho2020denoising}. Recent work has applied diffusion to language models~\cite{hoogeboom2021argmax}---generating text through iterative refinement over a denoising trajectory---as compared to standard next-token prediction language models~\cite{bachmann2024pitfalls}, achieving competitive performance with substantially higher throughput~\cite{li2022diffusion,song2025seed, odonoghue2026diffusiongemma, lin2026efficient}. DLMs broadly fall into three families: \textbf{continuous-space DLMs}, which diffuse over continuous embeddings before mapping back to discrete tokens~\cite{li2022diffusion, jo2026continuous}; \textbf{discrete-space DLMs}, which diffuse directly over tokens or vocabulary states via masking or transition matrices~\cite{ye2508dream, gong2025scaling}; and \textbf{hybrid AR--diffusion models}, which combine autoregressive generation with diffusion-based refinement~\cite{zhu2026flare}. We propose DLIG, an attribution method for interpreting DLM behavior. Although instantiated on a discrete DLM, our approach is broadly applicable to iterative diffusion processes.

\paragraph{Explainable AI (XAI), Attribution Methods and Diffusion Explainability.}
XAI encompasses post-hoc methods for explaining model decisions~\cite{linardatos2020explainable}, including SHAP~\cite{mosca2022shap}, LIME~\cite{ribeiro2016should}, and Integrated Gradients (IG)~\cite{sundararajan2017axiomatic}. IG quantifies feature contributions to a model's prediction and has been adapted to language models to measure the influence of individual input tokens on generated outputs~\cite{sanyal2021discretized}. Attention~\cite{vig2019multiscale} and probe-based~\cite{alain2016understanding} analyses can reveal information encoded by models and support hypothesis formation, but attribution methods more directly connect model behavior to specific inputs while remaining lightweight and interpretable~\cite{miglani2023using,aswal2025phonetic}. In diffusion models, post-hoc explanation has focused primarily on vision, either using diffusion to construct IG integration paths~\cite{kim2026diffusion, lei2024denoising} or deriving CAM-style visual attributions from denoising states in diffusion MLLMs~\cite{zuo2026diffusion}. For DLMs, however, post-hoc explanation remains largely conceptual: a recent taxonomy formalizes attribution contracts for iterative generation~\cite{nguyen2026attribution}, but no concrete method instantiates them. In particular, token-level attribution across layers and denoising steps remains open. We address this gap with DLIG, which extends IG to quantify a DLM's progressive commitment to its completion at arbitrary layers and denoising steps.

\paragraph{Mechanistic Interpretability.} Interpretability aims to reverse engineer the computations of machine learning models into human understandable algorithms~\cite{carvalho2019machine}. Mechanistic interpretability has emerged as a sub-domain that aims to interpret larger neural networks, typically large multimodal models~\cite{bereska2024mechanistic}. Standard techniques include observable methods such as training probes to decode linearly encoded information in the residual stream~\cite{elhage2021mathematical,alain2016understanding}, causality driven interventions that identify the causal role of internal representations through activation patching~\cite{zhang2024towards} and causal tracing~\cite{meng2022locating,aswal2026observable}, dictionary/concept-learning methods~\cite{shu2025survey,aswal2025conceptguard,bernas2026concepts,bhalla2026sparse}, and geometry and activation driven steering methods~\cite{fel2026structuring,wurgaft2026manifold}. Recent work has adapted these methods to interpret DLM behavior. \citet{shnaidman2025activation} extract a low-dimensional refusal direction from contrastive prompts and intervene on residual-stream activations during denoising,~\citet{an2026dlm} steer DLMs by biasing token distributions with pre-computed attribute scores,~\citet{li2026diffuguard} analyze safety across denoising steps and introduce representation-based auditing and repair,~\citet{yao2026timerome} adapt the original causal tracing technique for masked DLMs, and~\citet{kong2026should} use hidden-state geometry to identify diffusion-friendly layers for diffusion-based replacement. Rather than competing with interventional methods, DLIG provides a lightweight complementary first check: inexpensive attribution can identify or rule out candidate mechanisms before more costly causal validation.
\section{Diffusion Layer Integrated Gradients (DLIG)}
\label{sec:dlig}
Let a masked diffusion language model denoise a sequence over $T$ steps. At a fixed diffusion step $t$, let $X_t$ denote the partially-unmasked sequence and $C$ the conditioning prompt occupying positions $1,\dots,L$. We attribute a scalar score $F_t$ to the model's internal computation at a chosen layer $\ell$, isolating the causal contribution of the prompt to that score.

\paragraph{Scoring function.}
The score reads out the model's commitment to its own completion at step $t$. Let $z_t(\cdot)\in\mathbb{R}^{|V|}$ be the logits the model emits at each generated position, and let $y = (y_1,\dots,y_n)$ be the target tokens occupying the generated region (positions $L{+}1,\dots,L{+}n$). We use the summed target log-probability $F_t \;=\; \sum_{i=1}^{n} m_i \,\log \mathrm{softmax}\big(z_t^{(L+i)}\big)_{y_i},$ where $m_i\in\{0,1\}$ masks out positions not yet scored.\footnote{WiC (\S~\ref{sec:wic}) and infilling (\S~\ref{sec:rocstories}) have growing $n$, so we use a length-normalized, bounded \emph{mean-centered} $F_t$ instead; ProsQA (\S~\ref{sec:prosqa}) has step-invariant $n$ and uses the sum above.} By default $y$ is \emph{self-generated}: $y_i = (X_t)_{L+i}$, the tokens the model has itself committed to along its denoising trajectory, so the attribution targets actual model behavior. $F_t$ then measures how strongly the layer-$\ell$ activations drive the model toward the completion it is in the process of producing. We also implement the \emph{fixed-target} framing where $y$ is a supplied string held constant across prompts and diffusion steps, and $F_t$ scores the model's commitment to that string.

\paragraph{Layer activations and baseline.}
Let $a = h^{(\ell)}_t(X_t, C)$ be the layer-$\ell$ activations under the true prompt, and let $a' = h^{(\ell)}_t(X_t, \varnothing)$ be the activations under a null conditioning obtained by masking the prompt positions while holding the generated tokens $X_t$ fixed. The pair $(a',a)$ defines a straight-line path in activation space from the prompt-free to the prompt-present state.

\paragraph{Attribution.}
DLIG attributes $F_t$ to layer-$\ell$ activations by integrating the gradient of $F_t$ along that path and weighting by the displacement:
\[
\mathrm{DLIG}^{(\ell)}_t
\;=\;
(a - a')\;\odot\;
\int_{0}^{1}
\frac{\partial F_t}{\partial a}\Big( a' + \alpha\,(a - a') \Big)\, d\alpha,
\]
with $\odot$ the elementwise product, so that $\mathrm{DLIG}^{(\ell)}_t \in \mathbb{R}^{S\times H}$ carries one attribution value per (position, hidden-dim). The integral is approximated by an $m$-point Riemann sum,
\[
\mathrm{DLIG}^{(\ell)}_t
\;\approx\;
(a - a')\;\odot\;
\frac{1}{m}\sum_{k=1}^{m}
\frac{\partial F_t}{\partial a}\Big( a' + \tfrac{k}{m}\,(a - a') \Big).
\]

\paragraph{Per-Token Attribution.}
The tensor $\mathrm{DLIG}^{(\ell)}_t \in \mathbb{R}^{S\times H}$ carries a value per (position, hidden-dim). To score attributions, we collapse the hidden axis to one score per token, $s^{(\ell)}_t[i] = \sum_{j} \mathrm{DLIG}^{(\ell)}_t[i,j]$, and read a prompt token's attribution off at its position $i \le L$. Completeness (\S~\ref{sec:axioms}), $\sum_i s^{(\ell)}_t[i] = F_t(X_t, C) - F_t(X_t, \varnothing)$, makes the summed per-token attribution an \emph{exact additive decomposition} of the causal effect of inserting the prompt, with $s^{(\ell)}_t[i]$ being position $i$'s share of that total. Since the baseline masks the prompt, $s^{(\ell)}_t[i] > 0$ means the token drives the model toward its target, while $s^{(\ell)}_t[i] < 0$ means its presence \emph{suppresses} the target at layer $\ell$.

\paragraph{IG--DLIG Equivalence.}
Fix a layer $\ell$ and step $t$, and let $G_t^{(\ell)} : \mathbb{R}^{S\times H} \to \mathbb{R}$ denote the map that overrides the layer-$\ell$ output with a candidate activation $u$, recomputes the layers above $\ell$, and reads out the score:
$G_t^{(\ell)}(u) \;=\; F_t\big(\,\text{layers } \ell{+}1,\dots \text{ applied to } u\,\big).$
Then, DLIG is exactly IG applied to the function $G_t^{(\ell)}$ with input $a = h^{(\ell)}_t(X_t, C)$ and baseline
$a' = h^{(\ell)}_t(X_t, \varnothing)$:
\[
\mathrm{DLIG}^{(\ell)}_t[i,j]
\;=\;
\mathrm{IG}_{ij}\big(G_t^{(\ell)};\, a,\, a'\big)
\;=\;
(a - a')_{ij}
\int_{0}^{1}
\frac{\partial G_t^{(\ell)}}{\partial u_{ij}}
\big( a' + \alpha\,(a - a') \big)\, d\alpha .
\]

\subsection{Extending Axiomatic Properties of IG~\texorpdfstring{\cite{sundararajan2017axiomatic}}{[1]} to DLIG}
\label{sec:axioms}
The axiomatic guarantees of IG are proved for an arbitrary almost-everywhere-differentiable scalar function of the integration variable. Next, we demonstrate that these guarantees transfer to DLIG by instantiating that function as $G_t^{(\ell)}$, which is differentiable almost-everywhere as a composition of transformer blocks and a softmax readout, along with direct implementation tests where required~\footnote{Implementation checks (prerequisites for IG--DLIG correspondence to hold in practice) in Appendix~\ref{app:impl-checks}.}.

\textbf{(1) Completeness.}
The attribution summed over all coordinates equals the difference in the scored function between the endpoints of the integration path:
\[
\sum_{i,j} \mathrm{DLIG}^{(\ell)}_t[i,j]
= G_t^{(\ell)}(a) - G_t^{(\ell)}(a')
= F_t(X_t, C) - F_t(X_t, \varnothing).
\]
The identity holds for any scalar $F$ read off from $h^{(\ell)}$ up to the $O(1/m)$ error of the $m$-point Riemann sum, independent of how $F$ consumes the logits; it is therefore agnostic to the shifted-readout convention, which is checked separately below. Completeness also implies the Sensitivity(a) axiom---if $a$ and $a'$ differ only in coordinate $(i,j)$ and $G_t^{(\ell)}(a) \neq G_t^{(\ell)}(a')$, completeness applied to the pair forces $\mathrm{DLIG}^{(\ell)}_t[i,j] \neq 0$.

\textbf{(2) Implementation Invariance.}
The attribution depends on the network only through the function $G_t^{(\ell)}$ and its gradients. For any two parameterizations realizing the same map, $G_t^{(\ell)} \equiv \tilde{G}_t^{(\ell)}$ implies $\mathrm{DLIG}^{(\ell)}_t = \widetilde{\mathrm{DLIG}}^{(\ell)}_t$ at every $(a, a')$.

\textbf{(3) Linearity.}
For scalar weights $\lambda_1, \lambda_2$ and readouts $G_1, G_2$ of the same activation, $\mathrm{IG}(\lambda_1 G_1 + \lambda_2 G_2) = \lambda_1\,\mathrm{IG}(G_1) + \lambda_2\,\mathrm{IG}(G_2)$. Since $F_t$ enters DLIG only through $G_t^{(\ell)}$, attributions toward two fixed targets $y^{+}, y^{-}$ may be differenced,
$
\Delta\mathrm{DLIG}^{(\ell)}_t
\;=\;
\mathrm{DLIG}^{(\ell)}_t\big(y^{+}\big) - \mathrm{DLIG}^{(\ell)}_t\big(y^{-}\big)
\;=\;
\mathrm{IG}\big(G_t^{(\ell)}(\cdot\,;y^{+}) -
G_t^{(\ell)}(\cdot\,;y^{-});\, a,\, a'\big),
$
i.e.\ the difference is itself a valid path attribution of the contrast score $F_t(\cdot\,;y^{+}) - F_t(\cdot\,;y^{-})$, and satisfies completeness with respect to $\Delta F_t$.

\textbf{(4) Symmetry Preservation.}
DLIG integrates along the straight-line path $\gamma(\alpha) = a' + \alpha(a - a')$, the unique path choice for which the theorem holds; the property is inherited unchanged.
\section{Experimental Setup}
We demonstrate DLIG's utility in explaining DLM behavior across three tasks, all using DiffuGPT-M~\cite{gong2025scaling}, a masked diffusion language model adapted from GPT-2-medium~\cite{radford2019language}.\footnote{We use $T{=}64$ denoising steps and $m=12$ Riemann points (residual within tolerance, Appendix~\ref{app:impl-checks}).} Task-specific configurations, training details, and the per-task denoising-step grids used for attribution are in Appendix~\ref{app:model_training}.

\textbf{Word-Sense Disambiguation.} WiC~\cite{pilehvar2019wic} asks whether a target word shares its sense across two context sentences (binary). We wrap each instance in a prompt querying sense-match, with the model generating \texttt{Yes}/\texttt{No}; answer tokens share no vocabulary with the disambiguating context, avoiding surface-form overlap. As the official test labels are private, we re-split the train set into 4,928 train / 638 val / 500 test and finetune DiffuGPT-M.

\textbf{Multi-Hop Graph Reasoning.} ProsQA~\citep{hao2024training} is a synthetic multi-hop benchmark: each question states an ontology of typing facts (with off-path distractors), roots the queried subject, and poses a binary choice (``Is $s$ a $A$ or $B$?'') whose gold option is reachable by a $k$-hop chain. We finetune DiffuGPT-M on the 17,886 train / 300 val / 500 test split.

\textbf{Sentence Infilling.} ROCStories~\cite{mostafazadeh2016corpus} comprises five-sentence commonsense stories; we infill sentence three from the two preceding and two following sentences (1000 eval samples). The generated span attends bidirectionally to context on both sides, so DLIG can resolve how attribution mass splits between left and right context and evolves over denoising steps.

The main text reports each finding qualitatively; full quantitative results, effect sizes, and statistical significance testing for every experiment are in Appendix~\ref{app:stats}.

\begin{figure}[!ht]
    \centering
    \includegraphics[width=\linewidth]{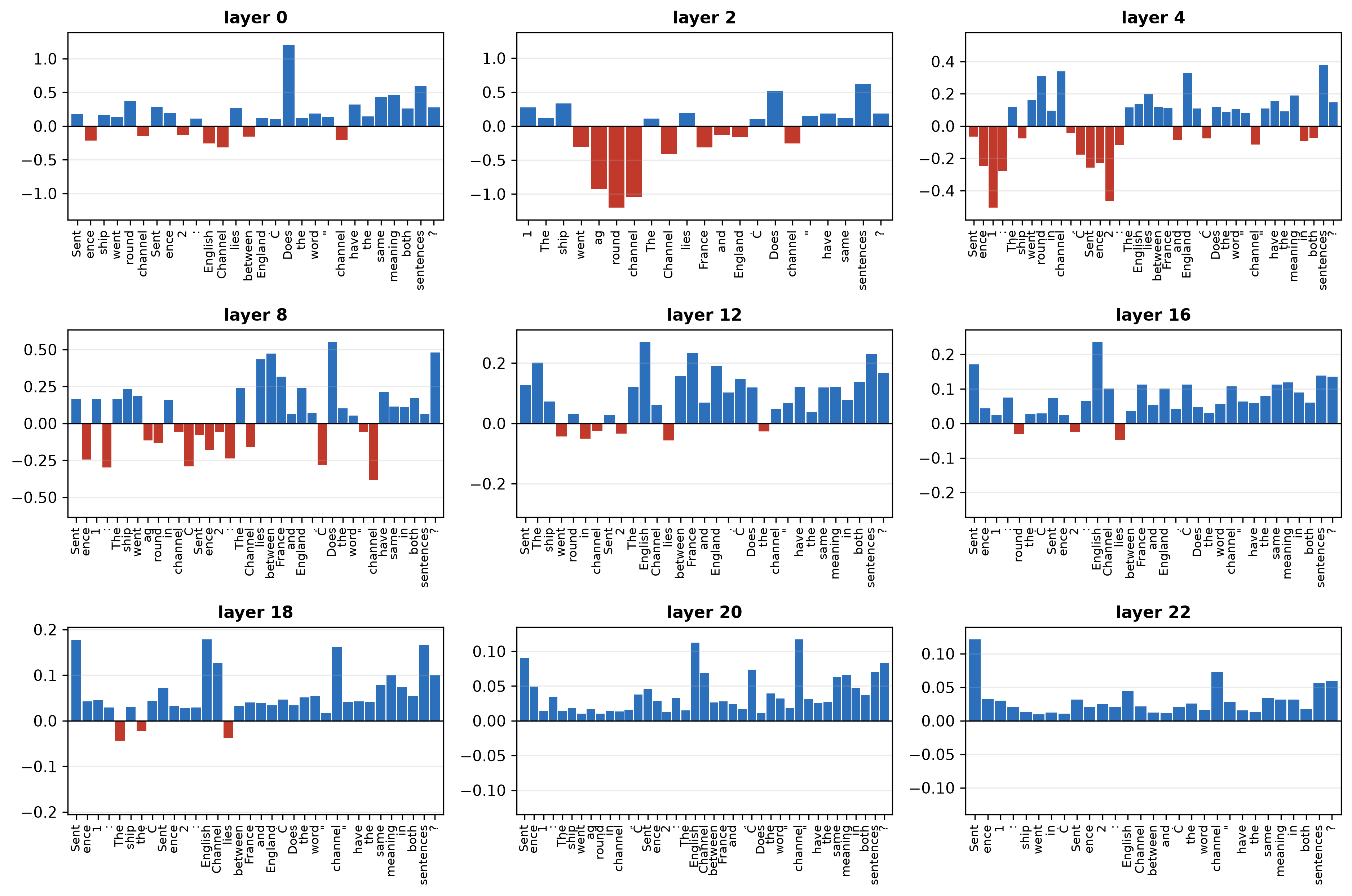}
    \caption{Layer-wise DLIG attribution for a single WiC example (target word \emph{channel}; gold label \emph{same senses}, model correctly answers \emph{Yes}). Bars are the per-token DLIG score averaged over all $T{=}64$ denoising steps. Blue (positive) marks tokens driving the model toward its committed answer, red (negative) those suppressing it. \textbf{\textit{Takeaway:}} Attribution concentrates on the two occurrences of the pivot word \emph{channel} and the surrounding content tokens that fix its shared sense, largest in the shallow layers and thinning with depth, while the prompt scaffolding stays near zero.}
    \label{fig:wic_attribution}
\end{figure}

\section{Word-Sense Disambiguation}
\label{sec:wic}
\paragraph{Setup.}
Scored under the self-generated framing of \S~\ref{sec:dlig}, the finetuned model is strongly \emph{Yes}-biased (class-wise performance in Appendix~\ref{app:model_training}, Table~\ref{tab:wic-training}). One possible explanation is that \textit{Yes} and \textit{No} decisions draw on prompt evidence at systematically different depths. DLIG provides an inexpensive test for this hypothesis directly from its layer-step attribution field. We contrast this attribution-based axis with a simple logit-based measure of when the model commits:

\begin{list}{$\bullet$}{
    \setlength{\leftmargin}{0.5em}
    \setlength{\labelwidth}{0.6em}
    \setlength{\labelsep}{0.2em}
    \setlength{\itemsep}{0.05em}
    \setlength{\topsep}{0.05em}
    \setlength{\parsep}{0pt}
    \setlength{\partopsep}{0pt}
}

\item \textbf{Attribution depth:} Per denoising step $t$, we weight each layer $\ell$ by its prompt-token attribution mass $M^{(\ell)}_t=\sum_{i\le L}|s^{(\ell)}_t[i]|$ and take the centroid $\bar{c}=\frac{1}{|\mathcal{T}'|}\sum_{t\in\mathcal{T}'}\big(\sum_\ell \ell\,M^{(\ell)}_t\big)/\big(\sum_\ell M^{(\ell)}_t\big)$, the mass-weighted mean layer index averaged over steps, measuring how deep the attribution sits.\footnote{Mass is zero at $22.3\%$ of steps (concentrated early, before the span commits; breakdown in Table~\ref{tab:wic-stepwise-zero-mass}, Appendix~\ref{app:stats}); these are excluded so the centroid isn't biased toward the input layer.}

\item \textbf{Commitment timing:} As a behavioral comparator, we directly record the softmax probability of the committed answer token across denoising steps and define the \textit{commitment step} as the earliest step from which it exceeds $0.5$ and remains above it.
\end{list}

We report each axis as a Mann--Whitney AUC~\cite{nachar2008mann} for \emph{Yes}- vs.\ \emph{No}-responses ($0.5$ chance), and additionally test each \emph{Yes}-response population for latent structure with Hartigan's dip test~\cite{hartigan1985dip}.

\begin{figure}[t]
  \centering
  \begin{subfigure}{0.5\linewidth}
    \includegraphics[width=\linewidth]{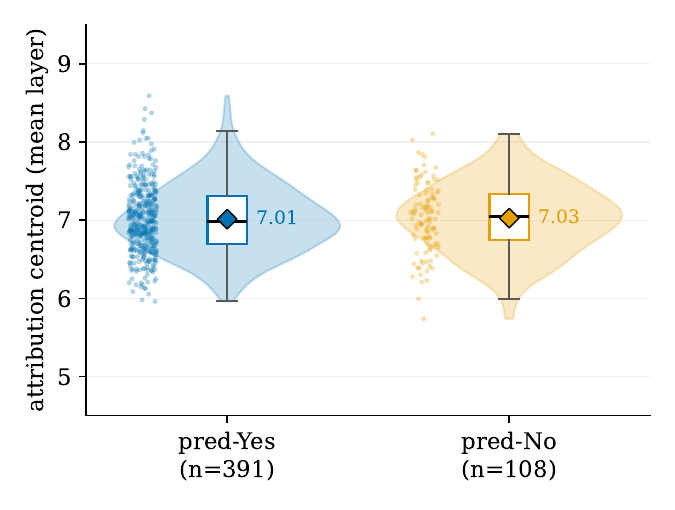}
    \caption{Attribution depth}\label{fig:wic-depth}
  \end{subfigure}\hfill
  \begin{subfigure}{0.5\linewidth}
    \includegraphics[width=\linewidth]{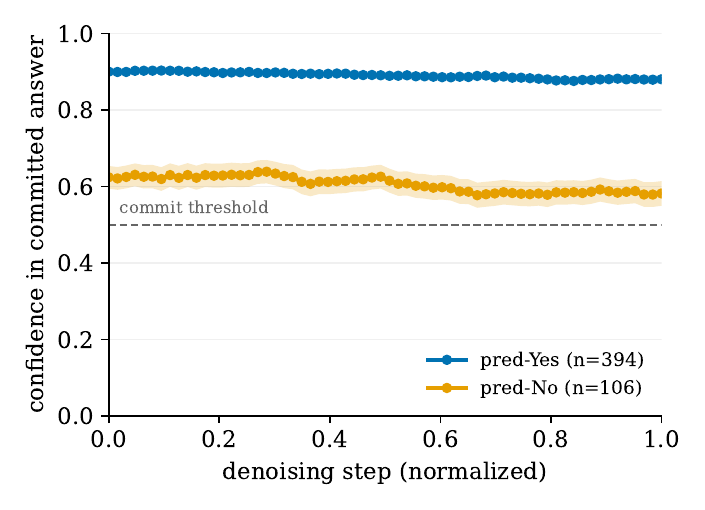}
    \caption{Commitment timing}\label{fig:wic-commit}
  \end{subfigure}
  \caption{Yes- vs.\ No-responses along the attribution-depth and commitment-timing axes, grouped by predicted label. \textbf{\textit{Takeaway:}} The two response types read out from essentially the same depth (a) but commit at different times (b), with Yes confident from the first step and No only barely clearing the threshold.}
  \label{fig:wic-axes}
\end{figure}

\paragraph{Results.}
We first illustrate in Figure~\ref{fig:wic_attribution} a single DLIG map on a correctly answered example~\footnote{Each layer's panel hides tokens with $|s^{(\ell)}_t[i]| < 0.08\max_i|s^{(\ell)}_t[i]|$ (keeping $\geq 3$ largest). Note that this is a hand-picked example chosen to demonstrate DLIG attribution on a clean case.}. The two occurrences of the pivot word \emph{channel} (``the channel'' and ``the English Channel'') together with the content tokens of both sentences carry the dominant, positive attribution, largest in the shallow layers which decreases with depth, while function words and template scaffolding stay near zero. The one early dissenting signal is \emph{France}: in the shallowest layers the model reads the geographic specificity of \emph{between France and England} as mild evidence \emph{against} the same-sense reading, but this is overridden with depth and flips to positive support by the mid layers, leaving the deep-layer map almost uniformly answer-aligned. Now, we analyze the attribution depth and commitment timing to ask whether the model's Yes- and No-responses, grouped by predicted label, are mechanically distinguishable along the two axes (parseable-prediction counts per axis in Figure~\ref{fig:wic-axes}).

\textbf{Yes- and No-responses read out from the same depth:} The two response types are distributed near-identically along this axis (Figure~\ref{fig:wic-depth}): both center just above layer~7 of the 24-layer stack, and the pred-Yes and pred-No distributions overlap almost entirely; DLIG provides cheap evidence against the depth hypothesis.

\textbf{Yes-responses commit stronger and hold it:} Reading the model's confidence in its own answer across the denoising trajectory (Figure~\ref{fig:wic-commit}) separates the two response types clearly: pred-Yes examples are already confident in Yes at the first denoising step and stay there, whereas pred-No examples begin only just above threshold and slightly degrade towards the end. The per-example commitment is markedly stronger for Yes than for No.

\section{Multi-Hop Graph Reasoning}
\label{sec:prosqa}

\paragraph{Setup.}
We analyze \textit{success} (correct or conceptually correct, $n{=}286$) \& \textit{failure} (incorrect option, $n{=}69$), excluding off-manifold generations ($n{=}145$).\footnote{Appendix~\ref{app:prosqa-bucket-examples} gives worked examples of each ProsQA bucket.} One hypothesis is that failures arise from answer-driving support being mislocalized onto distractors rather than the reasoning path; DLIG directly tests this over the prompt's fact graph. We construct each prompt's fact graph and mark two token sets: the \textbf{gold path} (BFS-recovered chain from subject to gold option)~\footnote{Hop distribution: $k{=}2{:}202$, $3{:}217$, $4{:}69$, $5{:}12$.}, and the \textbf{cited chain} (facts named in the model's generation and matched to graph facts via fuzzy string matching).

We define per-example targets ($y^{+}$) as the gold answer sentence and ($y^{-}$) as the same sentence with the gold option replaced by the incorrect choice, thereby isolating the answer concept.\footnote{$y^-$ is templated, not read off the model's text; Appendix~\ref{app:prosqa-bucket-examples} shows why it matches the model's answer on fails.} The per-token contrastive score $d[i]=s^{(\ell)}_t[i](y^{+})-s^{(\ell)}_t[i](y^{-})$ measures whether prompt token $i$ supports the correct $(d[i]>0)$ or incorrect $(d[i]<0)$ option. We define \textbf{path precision} $\mathrm{prec}=\sum_{i\in\mathrm{GOLD}}\max(d[i],0)\,/\,\sum_{i\in\mathrm{FACT}}\max(d[i],0)$ as the fraction of positive support attributable to gold-path tokens, measuring where positive support concentrates (facts or distractors). We report its gap over chance ($|\mathrm{GOLD}|/|\mathrm{FACT}|$) and against a random-edge null matched for path size.

Because $d$ tracks the gold option, its sign tracks model behavior only on successes; on fails, we therefore use $d_{\mathrm{behavior}}=-d$, so that $d_{\mathrm{behavior}}>0$ always marks support for the generated option. Along the reasoning chain, we average $d_{\mathrm{behavior}}$ per gold-path hop, normalizing within each example by its mean $|d_{\mathrm{behavior}}|$ across hops (so that examples contribute equally regardless of overall magnitude). We also report the \textbf{contrastive magnitude} $\overline{|d|}$, the mean $|d[i]|$ over fact tokens (total fact attention regardless of which option it supports).

\begin{figure}[t]
\centering
\begin{subfigure}[t]{0.32\linewidth}
    \centering
    \includegraphics[width=\linewidth]{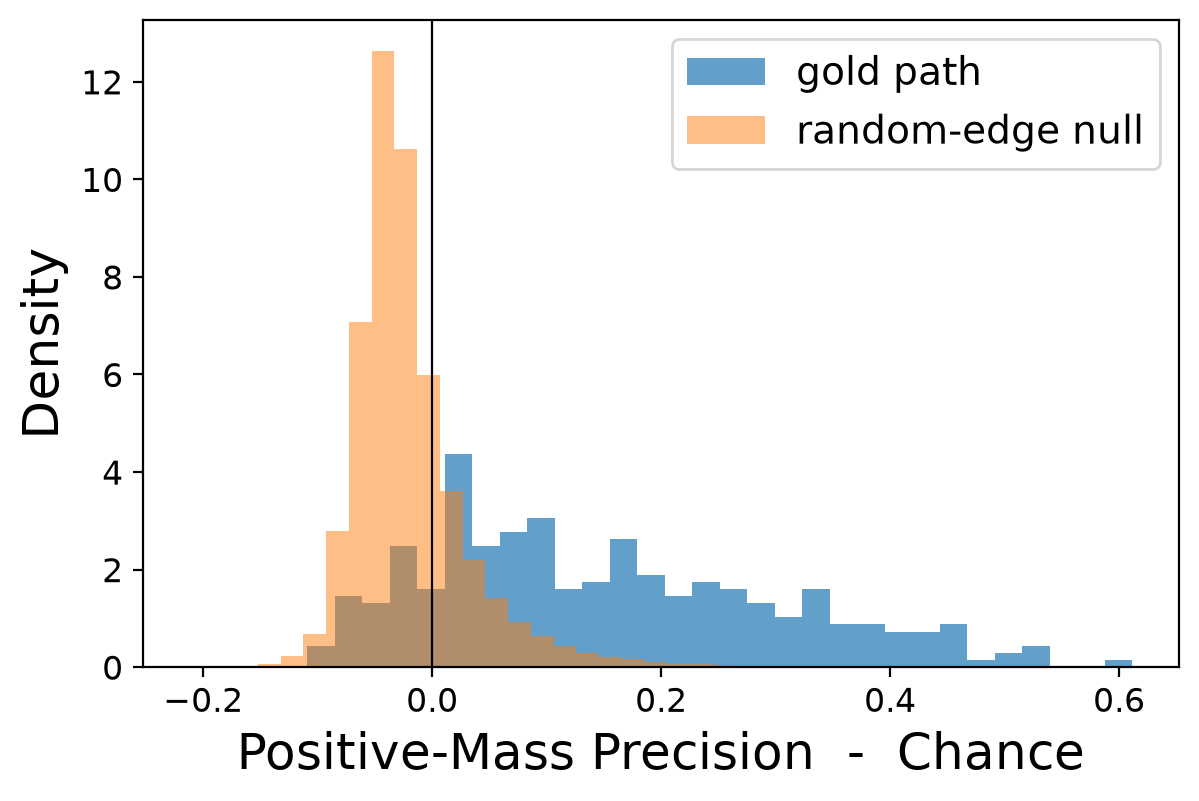}
    \caption{Positive contrastive mass on gold-path tokens versus a random-edge null, pooled over layers and denoising steps (successes).}
    \label{fig:prosqa-precision}
\end{subfigure}
\hfill
\begin{subfigure}[t]{0.32\linewidth}
    \centering
    \includegraphics[width=\linewidth]{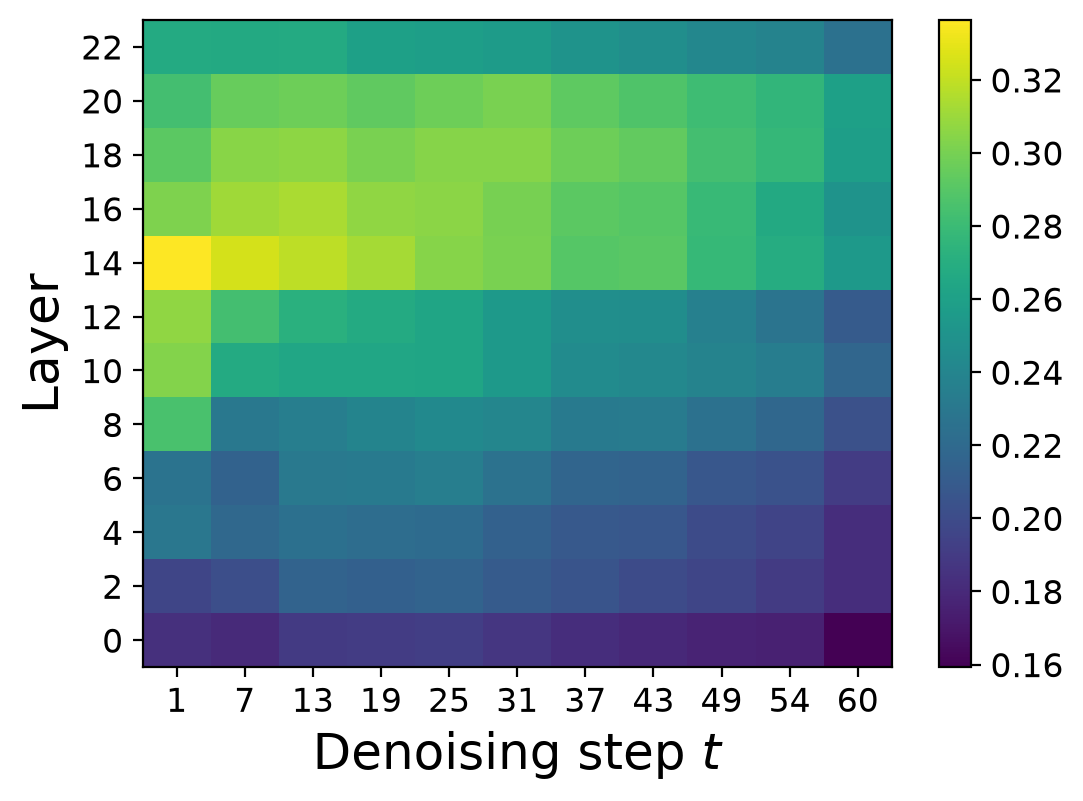}
    \caption{Gold-mass fraction by layer and denoising step.}
    \label{fig:prosqa-localization}
\end{subfigure}
\hfill
\begin{subfigure}[t]{0.32\linewidth}
    \centering
    \includegraphics[width=\linewidth]{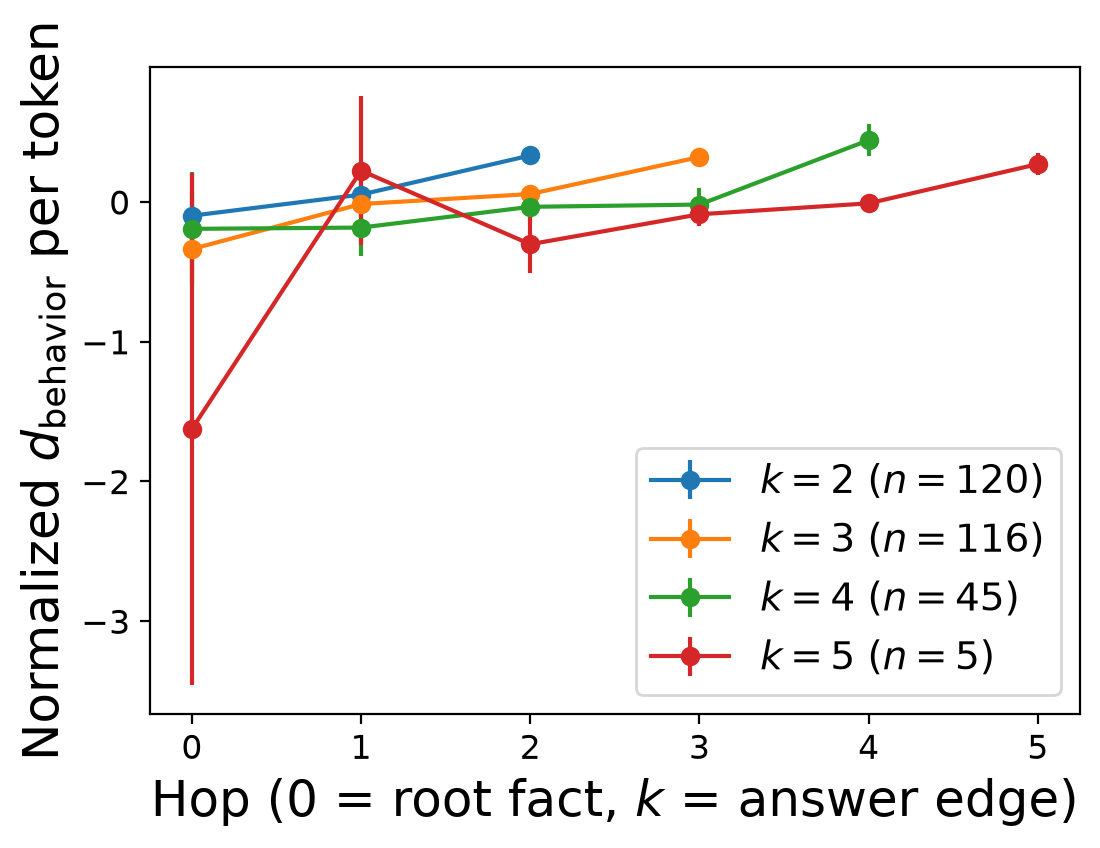}
    \caption{$d_{\mathrm{behavior}}$ per gold-path hop (0 = root fact, $k$ = answer edge), split by path length $k$; wide intervals at $k{=}5$ reflect the few length-5 chains in the test set.}
    \label{fig:prosqa-hop-profile}
\end{subfigure}

\caption{Attribution on the multi-hop reasoning graph (ProsQA, successes). \textbf{\textit{Takeaway:}} Positive contrastive mass concentrates on the gold reasoning path well above a random-edge null (a), localizes to the mid-to-upper layers and early denoising steps (b), and rises nearly monotonically along the chain toward the answer edge (c).}
\label{fig:prosqa-A-D}
\end{figure}

\paragraph{Results.}

\textbf{Attribution concentrates on answer driving facts:} $d_{\mathrm{behavior}}$'s positive mass concentrates on gold-path tokens well above chance and the random-edge null (Figure~\ref{fig:prosqa-precision}). Concentration is layer- and step-localized---rising towards later layers, decaying over denoising (Figure~\ref{fig:prosqa-localization})---and increases from the root fact to the answer edge at every path length (Figure~\ref{fig:prosqa-hop-profile}). On fails, attribution likewise concentrates on the model's cited chain rather than the remaining facts (Table~\ref{tab:prosqa-stats}); where cited and gold paths diverge, the two are indistinguishable.

\begin{figure}[t]
\centering
\includegraphics[width=\linewidth]{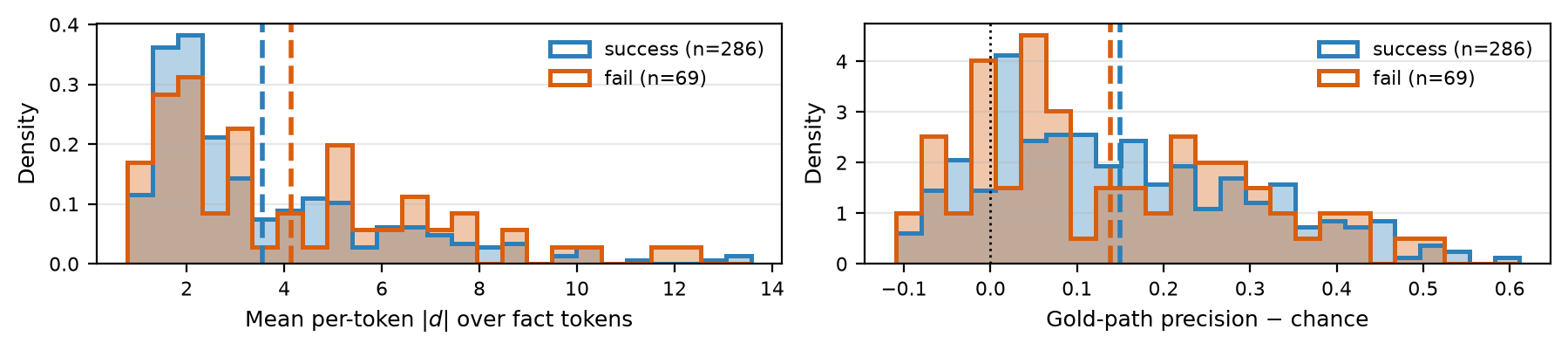}
\caption{Contrastive magnitude versus attribution location, by outcome. Left: mean per-token $|d|$ over fact tokens (how much support the facts carry). Right: gold-path precision gap (where that support lands). \textbf{\textit{Takeaway:}} Attribution localization is preserved across outcomes: failures localize the gold path as well as successes, while showing only a non-significant trend toward higher attribution magnitude.}
\label{fig:prosqa-magnitude-location}
\end{figure}

\textbf{Errors preserve attribution localization:} Gold-path precision is nearly identical for successes and fails and carries no signal in a joint logistic model (Figure~\ref{fig:prosqa-magnitude-location} right; Table~\ref{tab:prosqa-stats}). Fails show numerically higher contrastive magnitude $\overline{|d|}$, but this does not significantly discriminate outcomes overall or after difficulty controls. Thus, failures do not arise from attribution shifting onto distractors: DLIG shows that the relevant reasoning structure remains localized even when the model answers incorrectly.

\section{Sentence Infilling}
\label{sec:rocstories}

\paragraph{Setup.}
The model infills sentence 3 from sentences 1--2 (left) and 4--5 (right), scored under the self-generated framing of~\S~\ref{sec:dlig}. If successful infilling increasingly relies on context as the span resolves, this dependence should emerge over denoising; DLIG tests this directly. The position axis is signed distance to the span (negative = left, positive = right). Infill quality is final-span ROUGE-1 vs.\ gold.\footnote{For the same reason as in \S\ref{sec:wic}, steps with no scoreable committed span token carry $F_t\equiv0$ and are excluded, reducing the early-step sample (per-step sample sizes in Table~\ref{tab:infill-stepwise-mass}, Appendix~\ref{app:stats}).}

\paragraph{Results.}

\begin{figure}[t]
\centering
\begin{subfigure}[t]{0.40\linewidth}
    \centering
    \includegraphics[width=\linewidth]{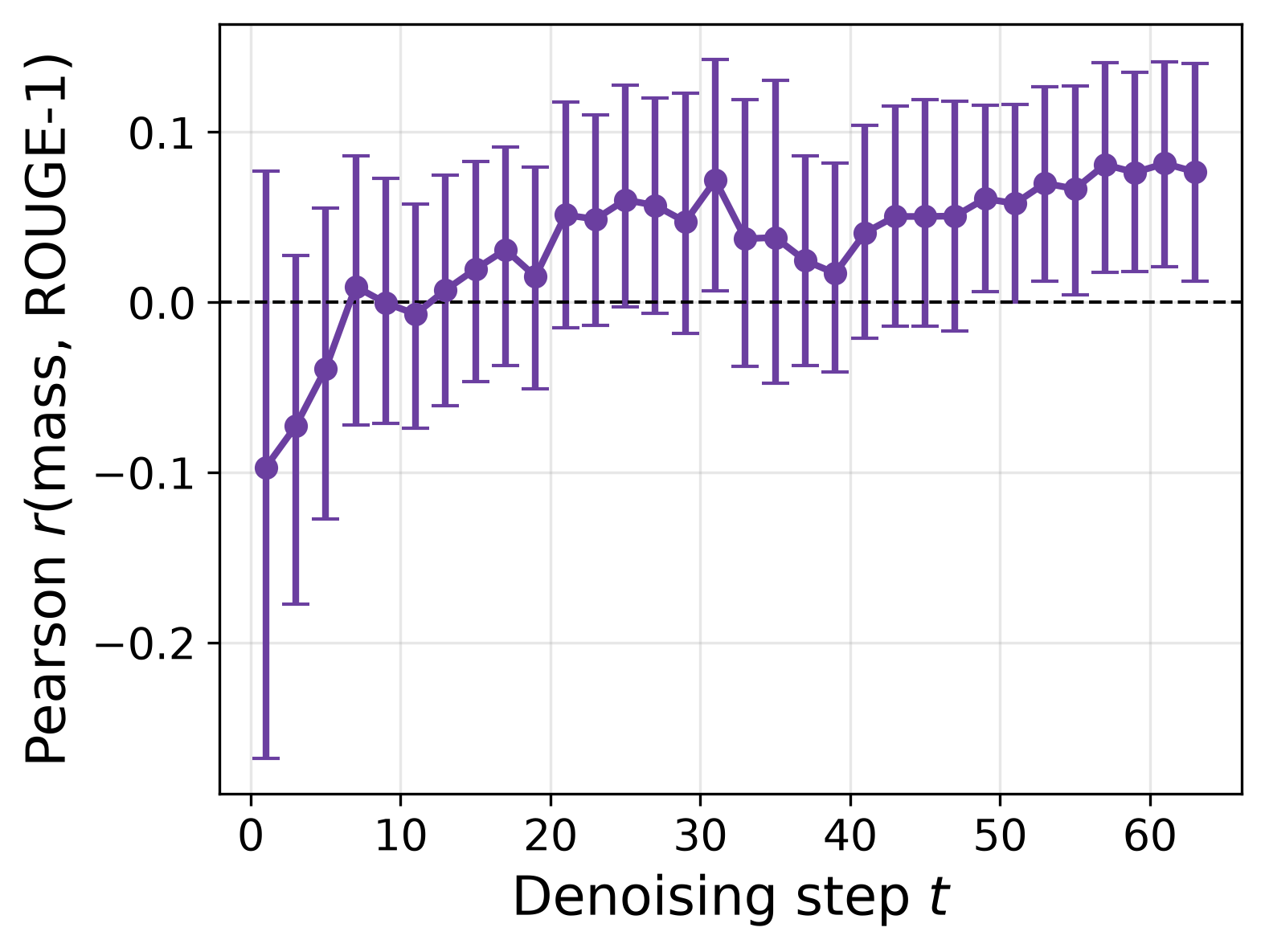}
    \caption{Per-step context-reliance vs.\ infill-quality correlation across denoising steps.}
    \label{fig:roc-rstep}
\end{subfigure}\hfill
\begin{subfigure}[t]{0.58\linewidth}
    \centering
    \includegraphics[width=\linewidth]{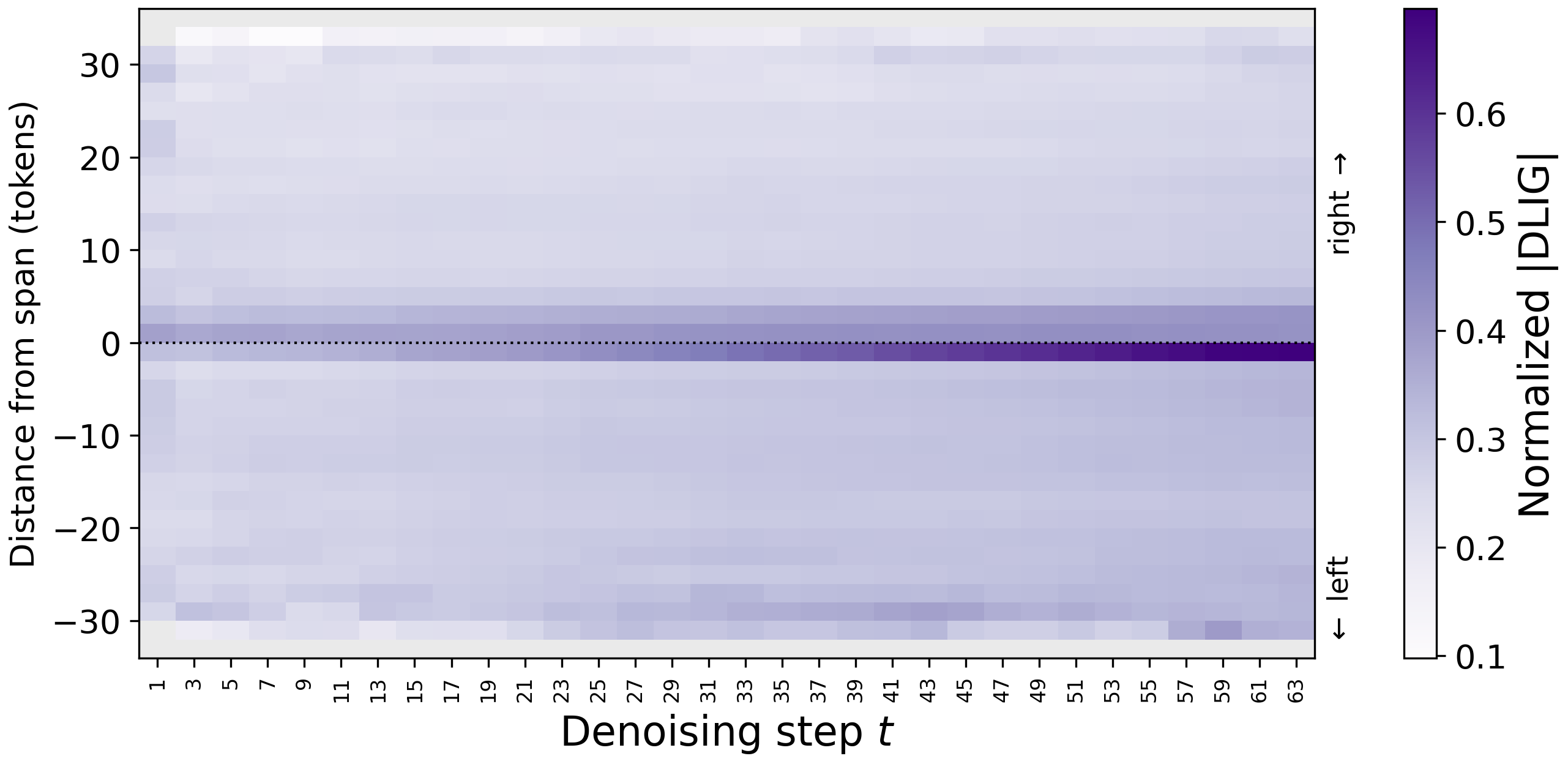}
    \caption{Signed-distance attribution profile per denoising step, split by low/high ROUGE.}
    \label{fig:roc-profile}
\end{subfigure}
\caption{Context reliance and attribution shape for sentence infilling (self-generated target). \textbf{\textit{Takeaway:}} The reliance--quality association is absent early and strengthens across denoising (\subref{fig:roc-rstep}); the attribution field is bidirectional and left-skewed, peaking just outside each span boundary, with better infills carrying slightly more mass throughout (\subref{fig:roc-profile}).}
\label{fig:roc-mass}
\end{figure}

\textbf{Context reliance predicts infill quality, increasingly over denoising:} The association between how much the model draws on context and how well it infills is absent at the first denoising steps and strengthens through the denoising (Figure~\ref{fig:roc-rstep}). ROUGE only scores the final span, evidencing that only once the model's committed span has sharpened toward that final output does the context become predictive of how well it reconstructs the span. Since there is no committed span early on, context reliance and eventual quality are unrelated. Pooling over steps hides this temporal structure (Appendix~\ref{app:roc-mass-auc}).

\textbf{Attribution is bidirectional, left-skewed, and strengthens late in denoising:} The model draws on both sides of the span, peaking at each boundary and decaying outward, with a persistently stronger left pull (a DLM sees right context an autoregressive model cannot; Figure~\ref{fig:roc-profile}). Context reliance grows as the model commits to its span later in denoising (Table~\ref{tab:infill-stepwise-mass}). Neither the amount nor the direction of context mass tracks infill quality (Appendix~\ref{app:roc-mass-auc}); the attribution field's shape is therefore task-level, while its late strengthening tracks commitment.

\section{Conclusion}
\label{sec:conclusion}

We introduce DLIG, a token-attribution method for DLMs that resolves a model's commitment to its output jointly across network depth and the denoising trajectory scored at any \texttt{(layer, step)}, decomposing it into per-token contributions. DLIG exposes not just where in the model support for an output lives, but how it evolves through its denoising trajectory.

We evaluate DLIG across three tasks, revealing structure a single-pass attribution would miss. On word-sense disambiguation, DLIG cheaply rules out attribution depth as an explanation for the model’s Yes bias: Yes and No responses draw on similar depths, despite sharply different commitment dynamics. On multi-hop reasoning, the DLM locates the right fact chain even when it answers wrongly, so errors trace to how it uses the facts and not where it looked. On sentence infilling, the DLM draws on both sides of the gap, including right context an autoregressive model cannot see, with context reliance correlating with infill quality only late in denoising.

Across tasks, DLIG is not only useful for visualizing attribution, but also serves to complement interventional analyses as an inexpensive first check for mechanistic hypotheses---identifying or ruling out \textit{where} and \textit{when} output-driving evidence emerges before more expensive interventional validation. Broadly, since a DLM arrives at its output gradually, explaining it means reading the whole denoising trajectory, \textbf{not} just its end. DLIG makes this \textit{dynamic} evolution of a DLM's commitment as well as where it draws from directly observable.

\section{Limitations and Future Work}
DLIG is a post-hoc gradient attribution method: it identifies where and when a model's committed output is gradient-sensitive to the prompt, not the causal circuit producing it, so its outputs are hypotheses to be confirmed, not mechanism at face value---future work should pair DLIG with interventional tests (activation patching, causal tracing~\cite{zhang2024towards,yao2026timerome}) against causal ground truth. We also instantiate DLIG on a single 355M discrete masked-diffusion model (DiffuGPT-M); extending to continuous-space and hybrid AR--diffusion families (\S~\ref{sec:background}), and to larger, natively-trained DLMs, is left for future work.

\clearpage
\bibliographystyle{unsrtnat}
\bibliography{main}
\appendix
\section{Extended Analyses}
\label{app:extended_analyses}
\subsection{Implementation Checks}
\label{app:impl-checks}
 
The checks below verify that the implementation realizes $G_t^{(\ell)}$ as defined in \S\ref{sec:axioms}, a precondition for the IG correspondence that no axiom can certify on its own.

\paragraph{Empirical Verification of Completeness.}
The verification tests three nested conditions at every (layer, step) pair used in the analyses. \textbf{(A) Hook transparency}: scoring $F_t$ through the intervention hook at the path endpoints $\alpha=0$ and $\alpha=1$ must reproduce a plain forward pass. \textbf{(B) Completeness}: the all-position DLIG sum is compared against $\Delta F = F_t(X_t,C) - F_t(X_t,\varnothing)$ under the combined tolerance $|\sum \mathrm{DLIG} - \Delta F| \le \varepsilon_{\mathrm{abs}} + \varepsilon_{\mathrm{rel}}\,|\Delta F|$ with $\varepsilon_{\mathrm{abs}}{=}5{\times}10^{-3}$, $\varepsilon_{\mathrm{rel}}{=}10^{-2}$; the absolute term is required because the Riemann discretization error is set by the curvature of $G_t^{(\ell)}$ along the path, not by $\Delta F$, so the relative error alone diverges as $\Delta F \to 0$. \textbf{(C) Convergence}: the relative error must decrease as $m$ grows (empirical signature of a correct integrator), checked at every grid cell between $m{=}200$ and $m{=}1000$. The check covers the full analyzed configuration: layers $\{0,2,\ldots,22\}$ $\times$ diffusion steps $\{1,3,5,7,9,11\}$ (72 cells) on a single trajectory with $T{=}12$, in \texttt{float32} under the completeness-validated (bounded, mean-centered) scoring readout on the finetuned checkpoint. The trajectory is seeded, so all cells share the same $X_t$ and the per-step score gaps are identical across layers.

\begin{figure}[h]
\centering
\includegraphics[width=0.85\linewidth]{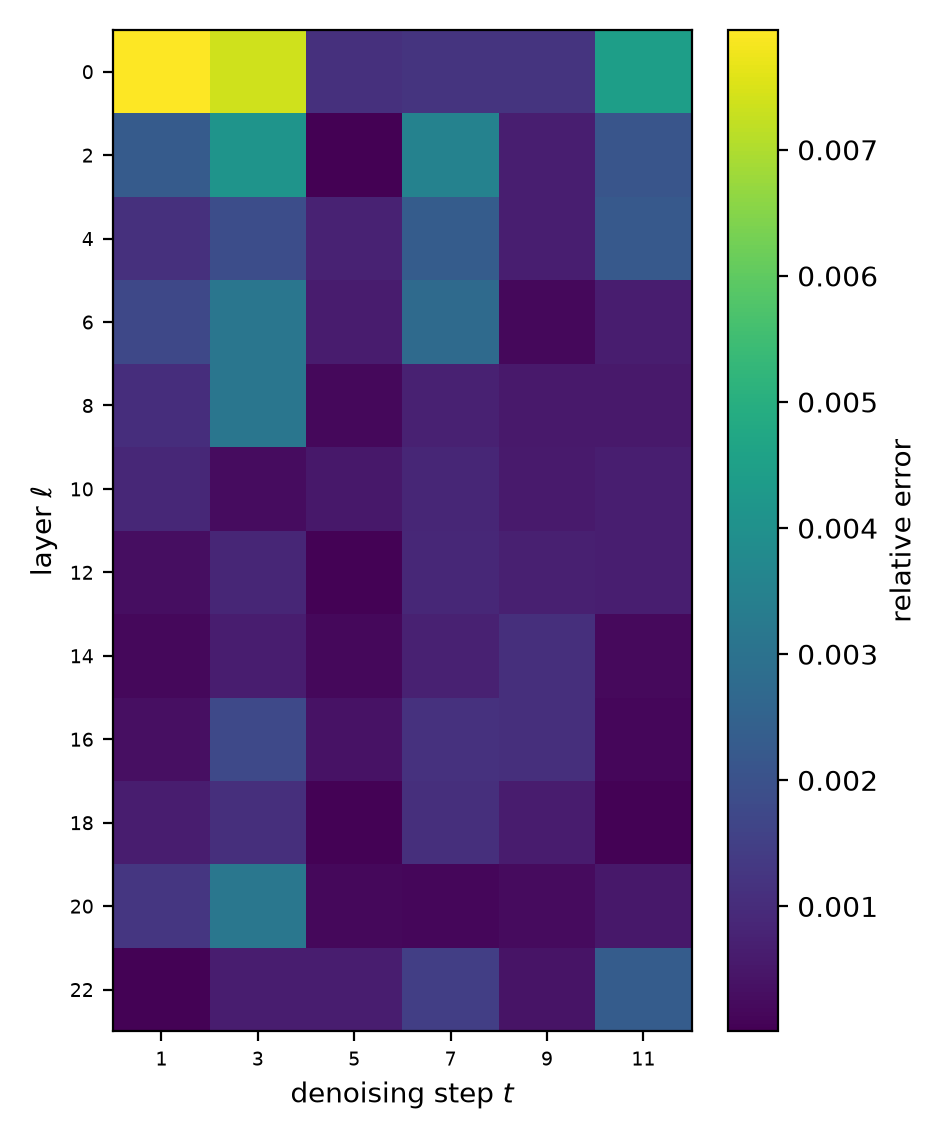}
\caption{Relative completeness error $|\sum\mathrm{DLIG}-\Delta F|/|\Delta F|$ at $m{=}1000$ over the full grid (layers $\times$ denoising steps). Every cell satisfies the combined tolerance $\varepsilon_{\mathrm{abs}}{+}\varepsilon_{\mathrm{rel}}|\Delta F|$; darker cells indicate smaller error.}
\label{fig:completeness-grid}
\end{figure}

\paragraph{Bidirectionality and Shift Gates.}
A \emph{bidirectionality} gate reads the first-block attention weights under an all-visible mask and checks that early positions place substantial mass on future positions; a masked diffusion LM must attend bidirectionally, and silently-causal attention would invalidate the attribution even while completeness still passes. A \emph{shift} gate confirms the declared \texttt{predicts\_shifted} convention by checking that, under that alignment, the argmax of the logits matches the committed tokens at unmasked generated positions; completeness is shift-agnostic and cannot detect a token-level off-by-one, so this is checked directly.

\paragraph{Partial-Forward Spot Check.}
DLIG uses a partial-forward optimization that resumes the network from the intervened layer rather than recomputing the layers below it. The spot check verifies exactness (layers below $\ell$ do not depend on the intervened activation) by comparing against the full-path computation on the real model, matching the per-position DLIG tensors produced by both.

\begin{figure}[h]
\centering
\includegraphics[width=0.7\linewidth]{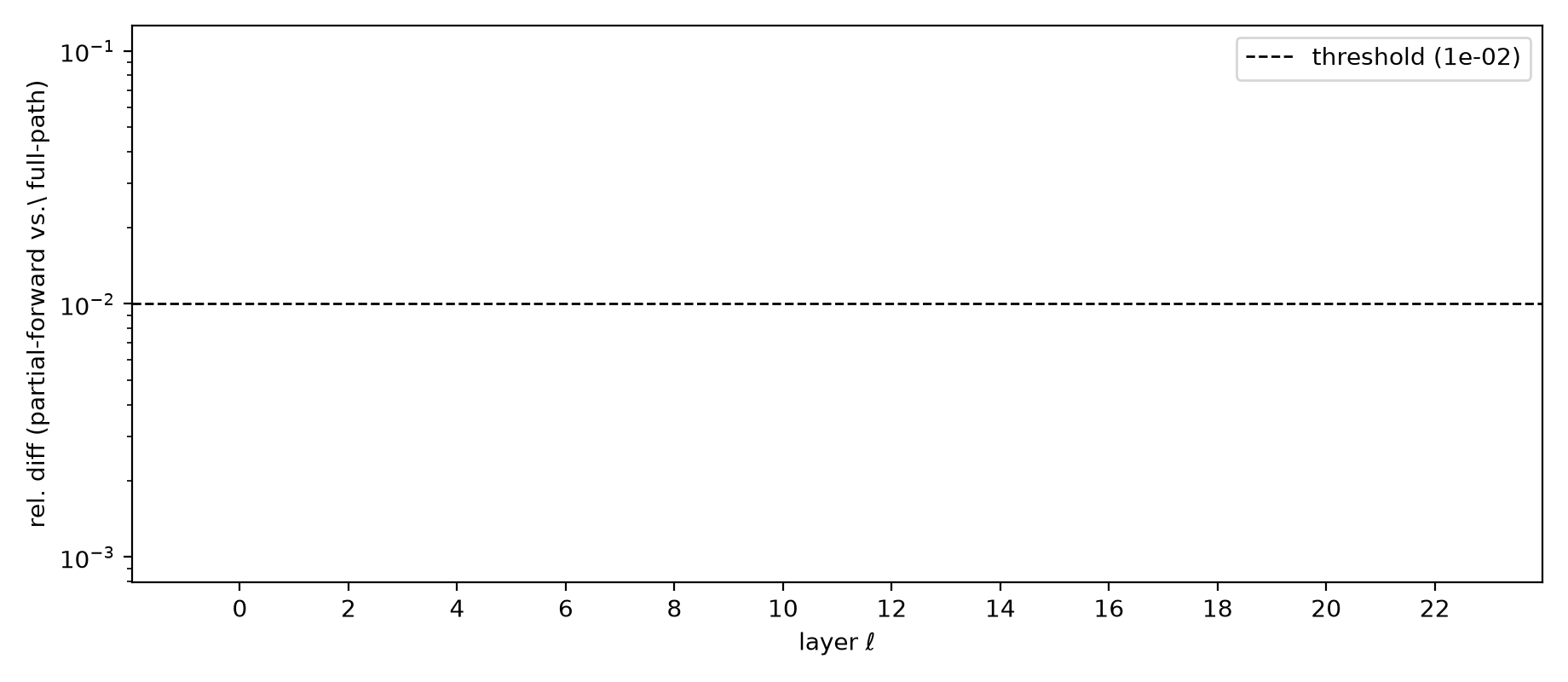}
\caption{Partial-forward vs.\ full-path DLIG, per layer: relative difference against the $10^{-2}$ pass/fail threshold. All layers agree exactly.}
\label{fig:spot-check-partial}
\end{figure}

\paragraph{Results.}
Endpoint scores recovered through the hook match the plain forward pass to floating-point precision at every grid cell, confirming hook transparency. At $m{=}1000$, every cell in the full completeness grid satisfies the combined tolerance (Figure~\ref{fig:completeness-grid}), with maximum absolute error $7.06\times10^{-3}$ and maximum relative error $7.95\times10^{-3}$, both at the $t{=}1$, $\ell{=}0$ cell; error is not uniform across the grid but concentrates at early denoising steps, where the score gap $\Delta F$ is smallest and the relative-error denominator is least forgiving, while absolute error stays within a comparable range throughout. Relative error decreases monotonically from $m{=}200$ to $m{=}1000$ at every cell, the expected signature of a converging Riemann integrator. The bidirectionality gate passes with forward-attention mass $0.924$ on the probed row, and the shift gate passes with an argmax--committed-token match rate of $0.789$ at unmasked generated positions. Finally, run at denoising step $t{=}2$ across the full analyzed layer set $\{0,2,4,\ldots,22\}$ on the base checkpoint, in the production \texttt{bfloat16} dtype with $m{=}12$ (Figure~\ref{fig:spot-check-partial}), the partial-forward and full-path DLIG computations agree exactly at every layer, with the maximum absolute and relative differences $0.00$ throughout. DiffuGPT-M bakes position information in once at the embedding layer rather than re-applying it per layer, so resuming from a cached activation is bit-exact rather than merely close, unlike RoPE-based backends where a small \texttt{bfloat16} roundoff floor is expected instead.

\subsection{Task Performance}
\begin{table}[!ht]
\centering
\small
\caption{Task-level performance of DiffuGPT-M on WiC, ProsQA, and ROCStories infilling. WiC: accuracy on all $500$ test examples, split by gold label into same-sense (Yes, $n{=}250$) and different-sense (No, $n{=}250$). ProsQA: outcomes over $n{=}500$ test examples, bucketed into \emph{success} (exact-match correct $\cup$ conceptually correct), \emph{fail} (wrong option among the two offered), and \emph{off-manifold} (wrong subject or an invalid/unlisted concept). Infill: ROUGE-1/2/L (self-generated span vs.\ gold sentence) over $n{=}1000$ ROCStories cases.}
\label{tab:task-performance}
\begin{tabular}{@{}llrl@{}}
\toprule
\textit{Task} & \textit{Outcome} & \textit{n} & \textit{\% [95\% CI]} \\
\midrule
\multirow[t]{3}{*}{WiC ($n{=}500$)}
& Overall accuracy & 318 & $63.6\ [59.2,\ 67.8]$ \\
& Same-sense (Yes) accuracy & 227 & $90.8\ [87.2,\ 94.4]$ \\
& Different-sense (No) accuracy & 91 & $36.4\ [30.4,\ 42.0]$ \\
\midrule
\multirow[t]{3}{*}{ProsQA ($n{=}500$)}
& Success (correct $\cup$ concept-only) & 275 & $55.0\ [51.0,\ 59.4]$ \\
& Fail (wrong option) & 64 & $12.8\ [10.0,\ 15.8]$ \\
& Off-manifold (wrong subject / invalid concept) & 161 & $32.2\ [28.2,\ 36.2]$ \\
\midrule
\multirow[t]{3}{*}{Infill ($n{=}1000$)}
& ROUGE-1 & \multicolumn{2}{c}{$12.02\ [11.22,\ 12.77]$} \\
& ROUGE-2 & \multicolumn{2}{c}{$1.22\ [0.93,\ 1.55]$} \\
& ROUGE-L & \multicolumn{2}{c}{$11.19\ [10.46,\ 11.86]$} \\
\bottomrule
\end{tabular}
\end{table}

Table~\ref{tab:task-performance} reports the performances on the Sentence Infilling task with the base DiffuGPT-M model, and the Word-Sense Disambiguation and Multi-Hop Graph Reasoning tasks with their respective finetuned variants of DiffuGPT-M. Note that these performances reflect the scores from a single evaluation run. The 95\% bootstrapped confidence intervals are calculated from 1,000 resamples within a single run. Due to the stochastic nature of the diffusion sampler, results should be expected to vary across runs.

\subsection{ProsQA Bucket and Contrastive-Target Examples}
\label{app:prosqa-bucket-examples}

Table~\ref{tab:prosqa-bucket-examples} gives one worked example per bucket (\S~\ref{sec:prosqa}): the question, the model's raw generation, and the resulting label. \emph{Success} is \emph{correct} $\cup$ \emph{concept\_only}; \emph{fail} is \emph{wrong\_valid}; \emph{subj\_wrong} and \emph{concept\_invalid} are the dropped off-manifold cases.

\begin{table}[h]
\centering
\small
\caption{One example per ProsQA bucket. \texttt{correct}: exact gold match. \texttt{concept\_only}: right subject and option, wrong wording. \texttt{wrong\_valid}: right subject, valid option (\emph{lempus}), just not gold. \texttt{subj\_wrong}: final sentence names the wrong subject (\emph{Every}, a parsing artifact, not a real entity). \texttt{concept\_invalid}: right subject, but the reasoning chain lands on a concept (\emph{numpus}) outside the offered pair \{worpus, chorpus\}.}
\label{tab:prosqa-bucket-examples}
\begin{tabularx}{\linewidth}{@{}l>{\raggedright\arraybackslash}Xl@{}}
\toprule
\textit{Bucket} & \textit{Model generation (\texttt{gen})} & \textit{Gold} \\
\midrule
correct & Bob is a yerpus. Every yerpus is a zumpusus. \#\#\# Bob is a shumpus. & Bob is a shumpus. \\
concept\_only & Bob is a gorpus. Every gorpus is a grimpus. Every grus is a a sterpus. Every sterpus is a numpus. \#\#\# \#\#\# Bob is a numpus. & Bob is a numpus. \\
wrong\_valid (fail) & Rex is a jelp. Every jelpus is a tumpus. Every tumpus is a wump. Every wumpus is a lempus. \#\#\# Rex is a real lempus. & Rex is a rompus. \\
subj\_wrong & Sally is a boompus. Every boompus is a remusus. Every remusus is a sterpus. Every Sally is a sterpus. & Sally is a sterpus. \\
concept\_invalid & Davis is a lorpus. Every lorpus is a numpus. \#\#\# Davis is a numpus. Every nump is a worpus. \#\#\# Davis is a worpus. & Davis is a worpus. \\
\bottomrule
\end{tabularx}
\end{table}

The \texttt{wrong\_valid} row above demonstrates the $y^-$ construction (\S~\ref{sec:prosqa}) concretely: the offered options are \{\emph{rompus} (gold), \emph{lempus}\}, so the dataset fixes wrong\_option${}=$\emph{lempus} independently of the model. The model's own generation ends ``\emph{Rex is a real lempus}'', i.e.\ its parsed \texttt{predicted\_option} is also \emph{lempus} --- the dataset-defined incorrect option and the model's committed answer coincide, exactly as guaranteed by the \texttt{wrong\_valid} gate. $y^+$ is then the gold sentence ``Rex is a rompus.'' and $y^-$ is the same template with the option swapped, ``Rex is a lempus.''

\subsection{ROCStories infilling: attribution profile and robustness}
\label{app:roc-fixed}

Here, we provide experiments to support \S~\ref{sec:rocstories} with (i) the pooled context-reliance vs.\ quality correlation, deferred here from the main text, and two robustness checks: (ii) the non-per-token normalization of context mass, and (iii) the fixed-target framing.

\begin{figure}[t]
\centering
\includegraphics[width=0.6\linewidth]{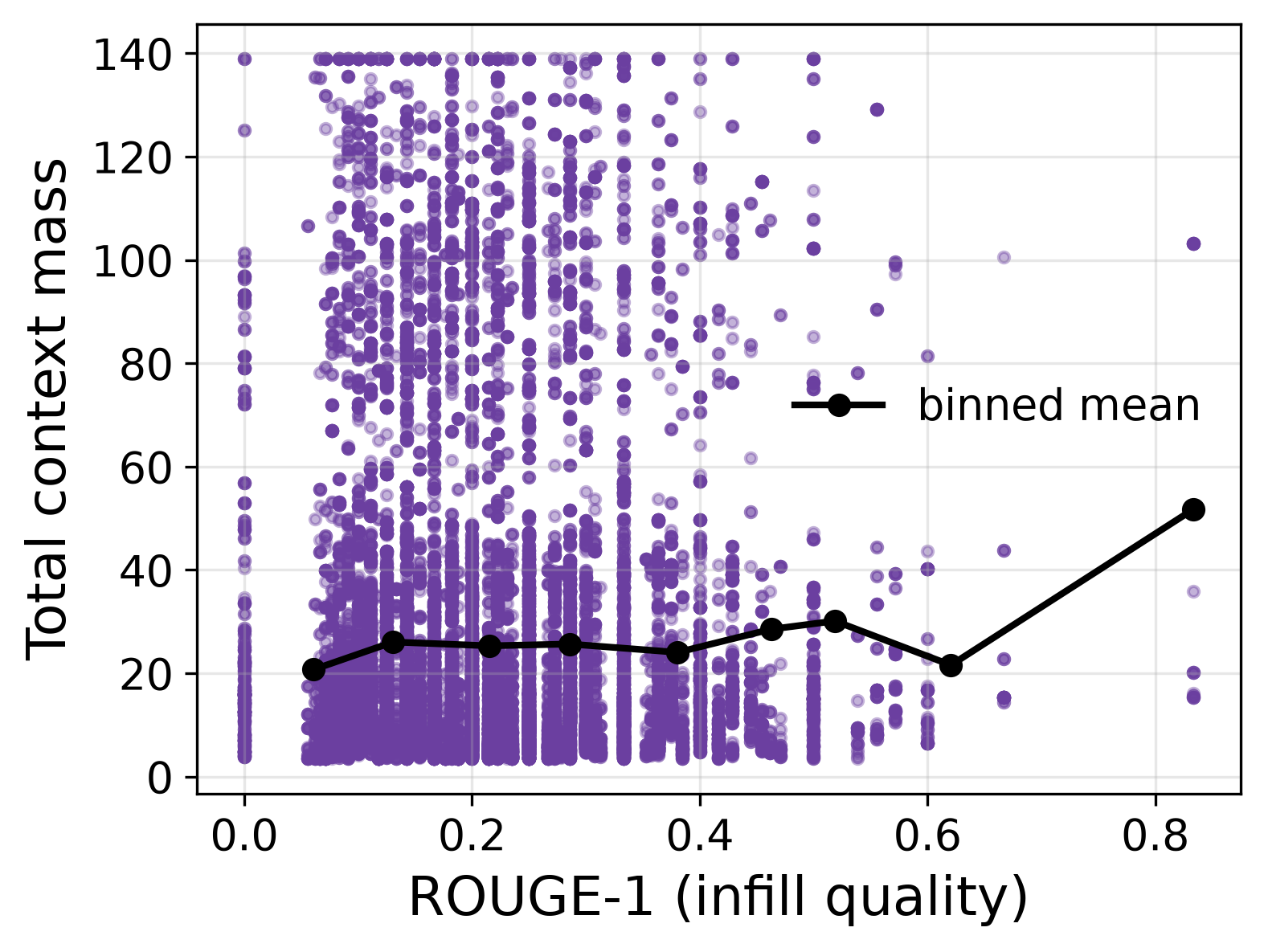}
\caption{Pooled over steps: total context mass vs.\ infill ROUGE-1 under the self-generated target (per-token normalized); the binned mean is roughly flat. The weak pooled correlation hides the trajectory that strengthens across denoising (main text, Figure~\ref{fig:roc-rstep}).}
\label{fig:roc-totalmass}
\end{figure}

\paragraph{Pooled context reliance vs.\ infill quality.}
\label{app:roc-pooled}
Figure~\ref{fig:roc-totalmass} shows total context mass against infill ROUGE pooled over the whole denoising trajectory. The pooled correlation is weak and its binned mean roughly flat, so no single-trajectory summary captures the reliance--quality relationship; only resolving the correlation per denoising step (main text, Figure~\ref{fig:roc-rstep}) reveals the association that emerges as the committed span converges toward the final output. Early steps additionally run on a reduced sample, as degenerate cells with no scoreable committed span token are excluded.

\paragraph{Context mass does not discriminate infill quality.}
\label{app:roc-mass-auc}
The main text states that the \emph{amount} of context mass does not track infill quality. We test this directly by splitting stories at the median ROUGE and reporting the Mann--Whitney AUC that a high-ROUGE story carries more total context mass than a low-ROUGE one, per denoising step over the full $32$-step trajectory so the claim is not obscured by pooling (pooled and per-step statistics in Tables~\ref{tab:roc-mass-auc-summary} and~\ref{tab:roc-mass-auc-per-step}). At no step does the AUC separate meaningfully from chance, and the right/left mass ratio is likewise flat across quality. Both the amount and the direction of context mass are therefore quality-neutral; only the per-step reliance--quality \emph{correlation} (Figure~\ref{fig:roc-rstep}) carries a quality signal, and only late in denoising.

\begin{table}[t]
\centering
\small
\caption{Context mass vs.\ infill quality, pooled statistics (self-generated target, median ROUGE split, $n_{\text{high}}{=}525$, $n_{\text{low}}{=}475$ stories). Per-step AUC in Table~\ref{tab:roc-mass-auc-per-step}.}
\label{tab:roc-mass-auc-summary}
\begin{tabularx}{\linewidth}{
    @{}
    >{\raggedright\arraybackslash}p{0.40\linewidth}
    >{\raggedright\arraybackslash}X
    >{\raggedleft\arraybackslash}p{0.18\linewidth}
    @{}
}
\toprule
\textit{Analysis} & \textit{Statistic} & \textit{Value} \\
\midrule
\multirow[t]{2}{=}{Total mass, high- vs.\ low-ROUGE (pooled; pseudo-replicated, see per-step)}
& AUC [95\% CI] ($n_{\text{high}}{=}14{,}966$, $n_{\text{low}}{=}13{,}548$ story-step cells)
& $0.510\;[0.503,0.517]$ \\
& $p$
& $3.4{\times}10^{-3}\;^{**}$ \\
\midrule
\multirow[t]{2}{=}{Right/left mass ratio vs.\ ROUGE-1}
& Pearson $r$ ($n{=}28{,}514$)
& $0.027$ \\
& Mean ratio (right/left)
& $0.984$ \\
\bottomrule
\end{tabularx}
\end{table}

\begin{table}[t]
\centering
\small
\caption{Mann--Whitney AUC that a high-ROUGE story carries more total context mass than a low-ROUGE one, per denoising step (self-generated target, median ROUGE split, $n_{\text{high}}{=}525$, $n_{\text{low}}{=}475$ stories; $0.5{=}$no difference). $95\%$ bootstrap CIs over stories. No step reaches significance ($p{>}0.07$ throughout).}
\label{tab:roc-mass-auc-per-step}
\begin{tabular}{@{}rlr@{}}
\toprule
$t$ & AUC [95\% CI] & $p$ \\
\midrule
1  & 0.501 [0.401, 0.606] & 0.986 \\
3  & 0.500 [0.441, 0.560] & 0.989 \\
5  & 0.485 [0.435, 0.536] & 0.565 \\
7  & 0.498 [0.456, 0.545] & 0.938 \\
9  & 0.504 [0.462, 0.547] & 0.849 \\
11 & 0.490 [0.450, 0.530] & 0.620 \\
13 & 0.503 [0.464, 0.543] & 0.879 \\
15 & 0.501 [0.463, 0.537] & 0.945 \\
17 & 0.517 [0.479, 0.553] & 0.364 \\
19 & 0.504 [0.469, 0.542] & 0.816 \\
21 & 0.517 [0.479, 0.554] & 0.356 \\
23 & 0.517 [0.481, 0.553] & 0.350 \\
25 & 0.524 [0.489, 0.558] & 0.202 \\
27 & 0.529 [0.493, 0.563] & 0.123 \\
29 & 0.529 [0.493, 0.563] & 0.118 \\
31 & 0.532 [0.494, 0.568] & 0.086 \\
33 & 0.533 [0.498, 0.570] & 0.076 \\
35 & 0.529 [0.491, 0.565] & 0.120 \\
37 & 0.516 [0.481, 0.551] & 0.396 \\
39 & 0.514 [0.477, 0.549] & 0.444 \\
41 & 0.521 [0.486, 0.554] & 0.248 \\
43 & 0.523 [0.486, 0.557] & 0.211 \\
45 & 0.514 [0.478, 0.551] & 0.436 \\
47 & 0.514 [0.482, 0.549] & 0.432 \\
49 & 0.507 [0.470, 0.543] & 0.708 \\
51 & 0.500 [0.464, 0.534] & 0.992 \\
53 & 0.499 [0.463, 0.536] & 0.950 \\
55 & 0.491 [0.454, 0.526] & 0.634 \\
57 & 0.501 [0.463, 0.537] & 0.965 \\
59 & 0.490 [0.455, 0.524] & 0.573 \\
61 & 0.491 [0.455, 0.525] & 0.629 \\
63 & 0.486 [0.451, 0.522] & 0.443 \\
\bottomrule
\end{tabular}
\end{table}

\begin{figure}[t]
\centering
\includegraphics[width=\linewidth]{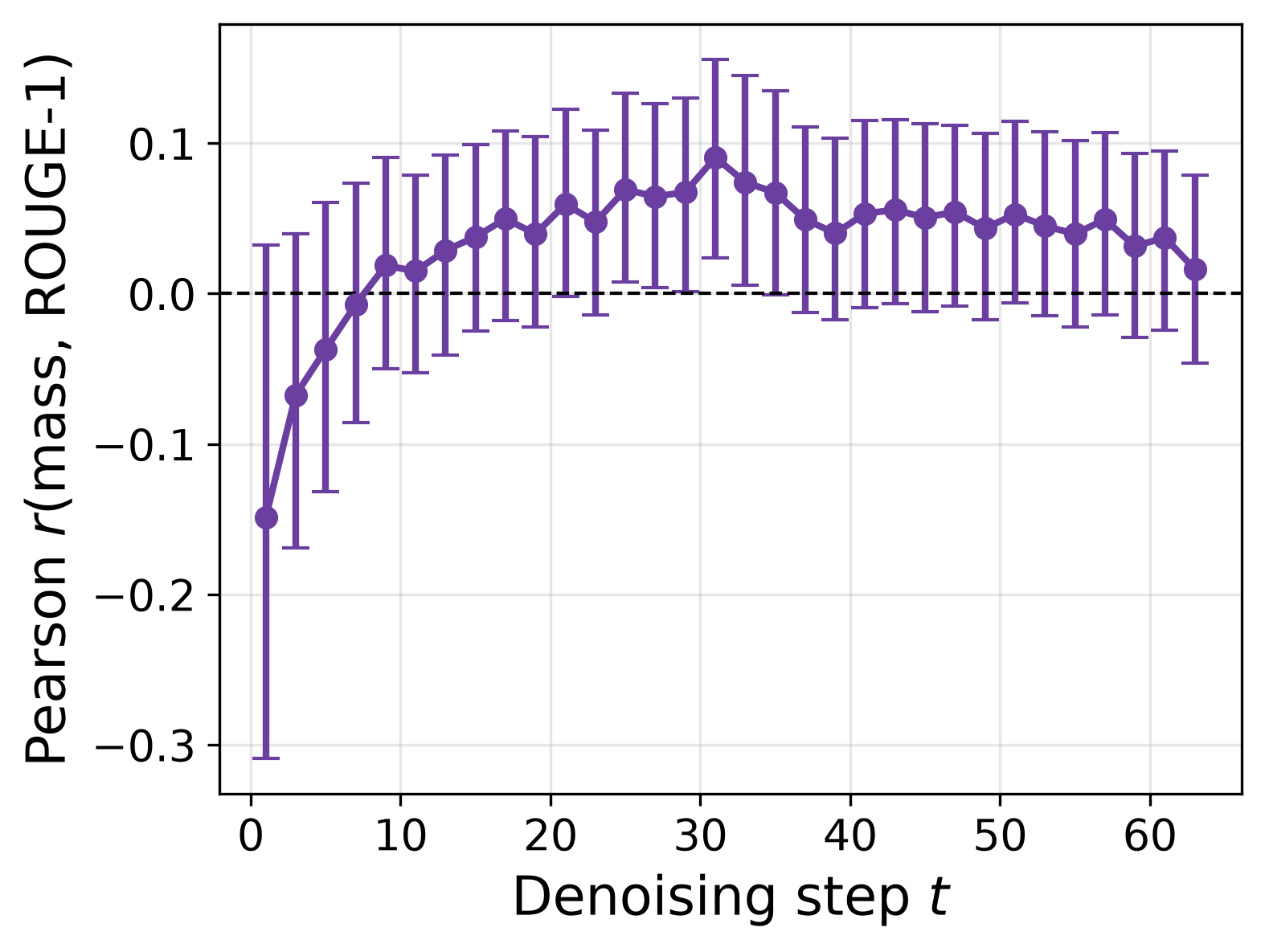}
\caption{Normalization robustness: per-step Pearson $r$(mass, ROUGE-1) for the self-generated target \emph{without} per-token normalization ($95\%$ bootstrap CI). The near-zero-early, strengthening-over-denoising pattern of Fig.~\ref{fig:roc-rstep} survives; per-token normalization only sharpens the late-step peak.}
\label{fig:roc-rstep-raw}
\end{figure}

\paragraph{Context mass without per-token normalization.}
\label{app:roc-normalization}
The main-text total context-mass results use per-token normalization (mass divided by the number of scoreable target tokens) to remove the growing-target-size trend intrinsic to the self-generated framing. Figure~\ref{fig:roc-rstep-raw} shows the per-step correlation without normalization: the qualitative pattern is unchanged, a correlation near zero early that strengthens over denoising, confirming the trend is not a normalization artifact. Per-token normalization simply sharpens the late-step signal.

\subsubsection{Fixed-target framing}
\label{app:roc-fixed-target}

Table~\ref{tab:roc-gold-fixed-target-stats} reports the full statistics for this robustness check (fixed gold target, $y$ = gold sentence~3 held constant across steps; \S~\ref{sec:rocstories}).

\begin{table}[t]
\centering
\small
\caption{Fixed-target (gold) robustness statistics for ROCStories infilling (\S~\ref{app:roc-fixed-target}), $n{=}1000$ stories $\times$ 32 denoising steps.}
\label{tab:roc-gold-fixed-target-stats}
\begin{tabularx}{\linewidth}{
    @{}
    >{\raggedright\arraybackslash}p{0.40\linewidth}
    >{\raggedright\arraybackslash}X
    >{\raggedleft\arraybackslash}p{0.18\linewidth}
    @{}
}
\toprule
\textit{Analysis} & \textit{Statistic} & \textit{Value} \\
\midrule
\multirow[t]{3}{=}{Total mass vs.\ ROUGE-1 (per-token normalized)}
& Pooled Pearson $r$ ($n{=}32{,}000$)
& $0.197$ \\
& Per-step $r$ range (32 steps, all CIs clear of $0$)
& $[0.192,\ 0.241]$ \\
& Self-generated pooled $r$, for comparison (\S~\ref{sec:rocstories})
& $0.013$ \\
\midrule
\multirow[t]{2}{=}{Left-tail attribution magnitude}
& Mean $|d|$ at $t{=}1$ vs.\ $t{=}63$
& $2.4$ / $1.0$ \\
\midrule
\multirow[t]{3}{=}{Right/left mass ratio vs.\ ROUGE-1}
& Pearson $r$ ($n{=}32{,}000$)
& $0.038$ \\
& Mean ratio (right/left), fixed target
& $1.121$ \\
& Mean ratio (right/left), self-generated, for comparison
& $0.969$ \\
\bottomrule
\end{tabularx}
\end{table}

\begin{figure}[t]
\centering
\begin{subfigure}[t]{0.48\linewidth}
    \centering
    \includegraphics[width=\linewidth]{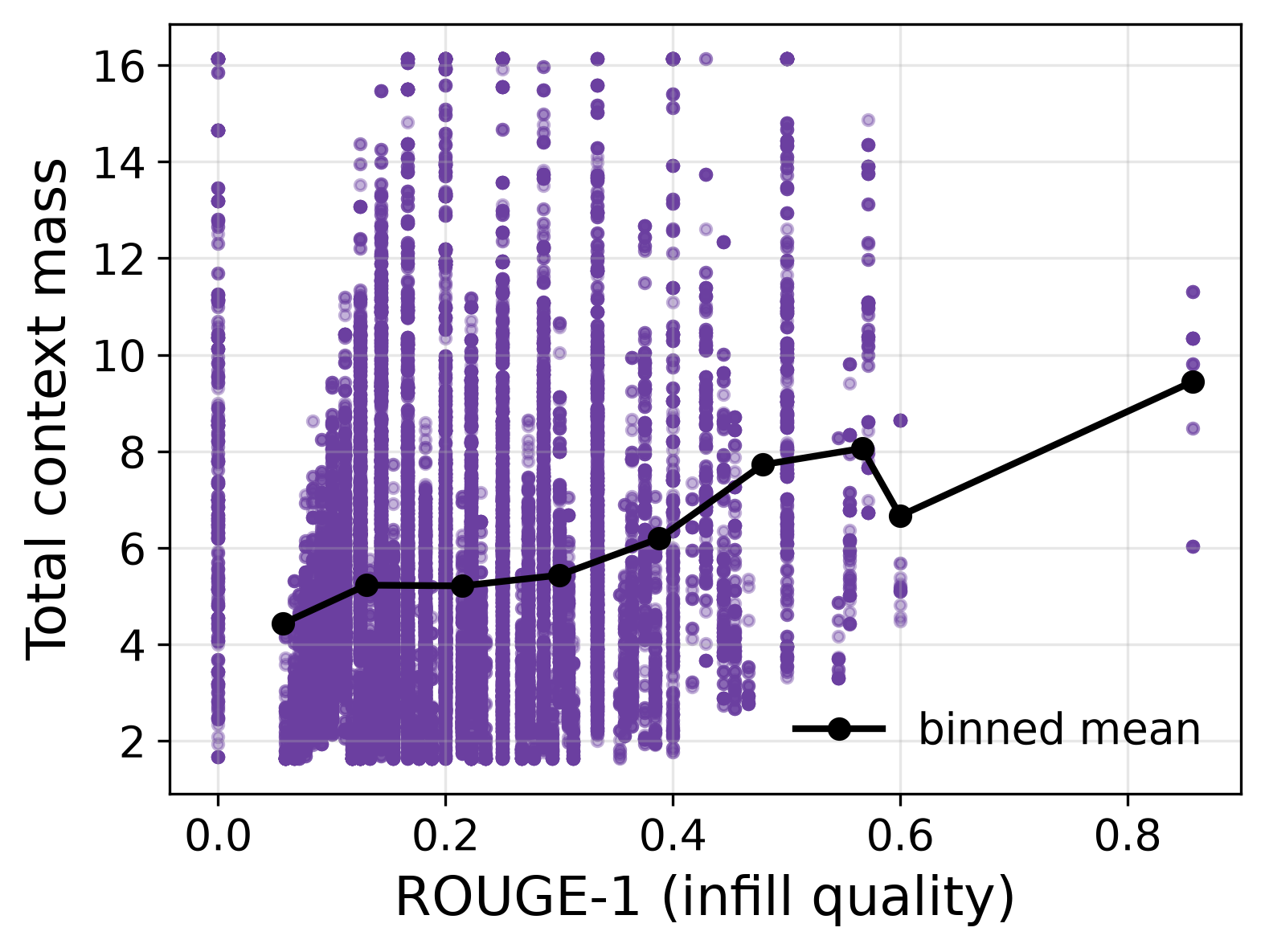}
    \caption{Pooled over steps: total context mass vs.\ ROUGE-1 --- substantially larger than the self-generated value, for the non-causal reason discussed in the text.}
    \label{fig:roc-totalmass-gold}
\end{subfigure}\hfill
\begin{subfigure}[t]{0.48\linewidth}
    \centering
    \includegraphics[width=\linewidth]{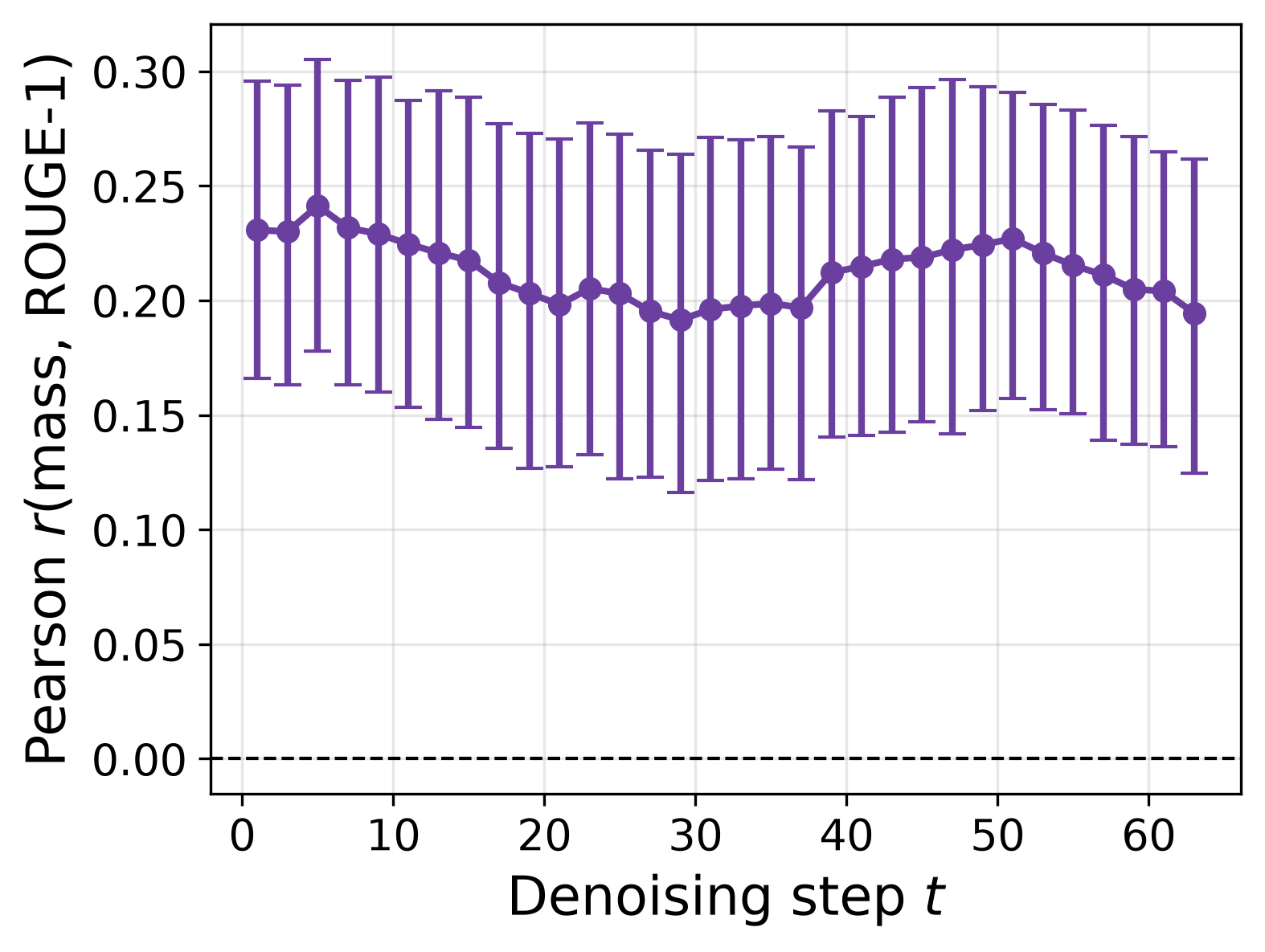}
    \caption{Per-step Pearson $r$ ($95\%$ bootstrap CI): elevated across all 32 denoising steps, with no over-denoising rise, in contrast to the self-generated trajectory (Fig.~\ref{fig:roc-rstep}).}
    \label{fig:roc-rstep-gold}
\end{subfigure}
\caption{Fixed-target robustness: context reliance vs.\ infill quality under the fixed gold target (per-token normalized). The correlation is an order of magnitude larger than the self-generated value and, unlike it, does not rise over denoising.}
\label{fig:roc-mass-gold}
\end{figure}

\paragraph{Fixed-target framing inflates the reliance--quality correlation.}
Under the fixed target, total context mass correlates with infill quality far more strongly than under the self-generated framing (Fig.~\ref{fig:roc-totalmass-gold}), and the per-step correlation stays large and roughly flat across the full trajectory rather than rising late, as it does under the self-generated framing (Fig.~\ref{fig:roc-rstep-gold}). We hypothesize that since ROUGE measures overlap between the generation and the gold sentence, on high-ROUGE stories the fixed gold target lies close to the model's actual trajectory and receives elevated gradient signal for that reason alone, independent of how much the model relied on context. Fixing the target thus folds a target-proximity component into the measured correlation---present from the first step---which the self-generated framing removes, reflecting the behavior-anchored correlation that survives the framing change based on the model's use of context in producing its own output.

\begin{figure}[t]
\centering
\includegraphics[width=0.8\linewidth]{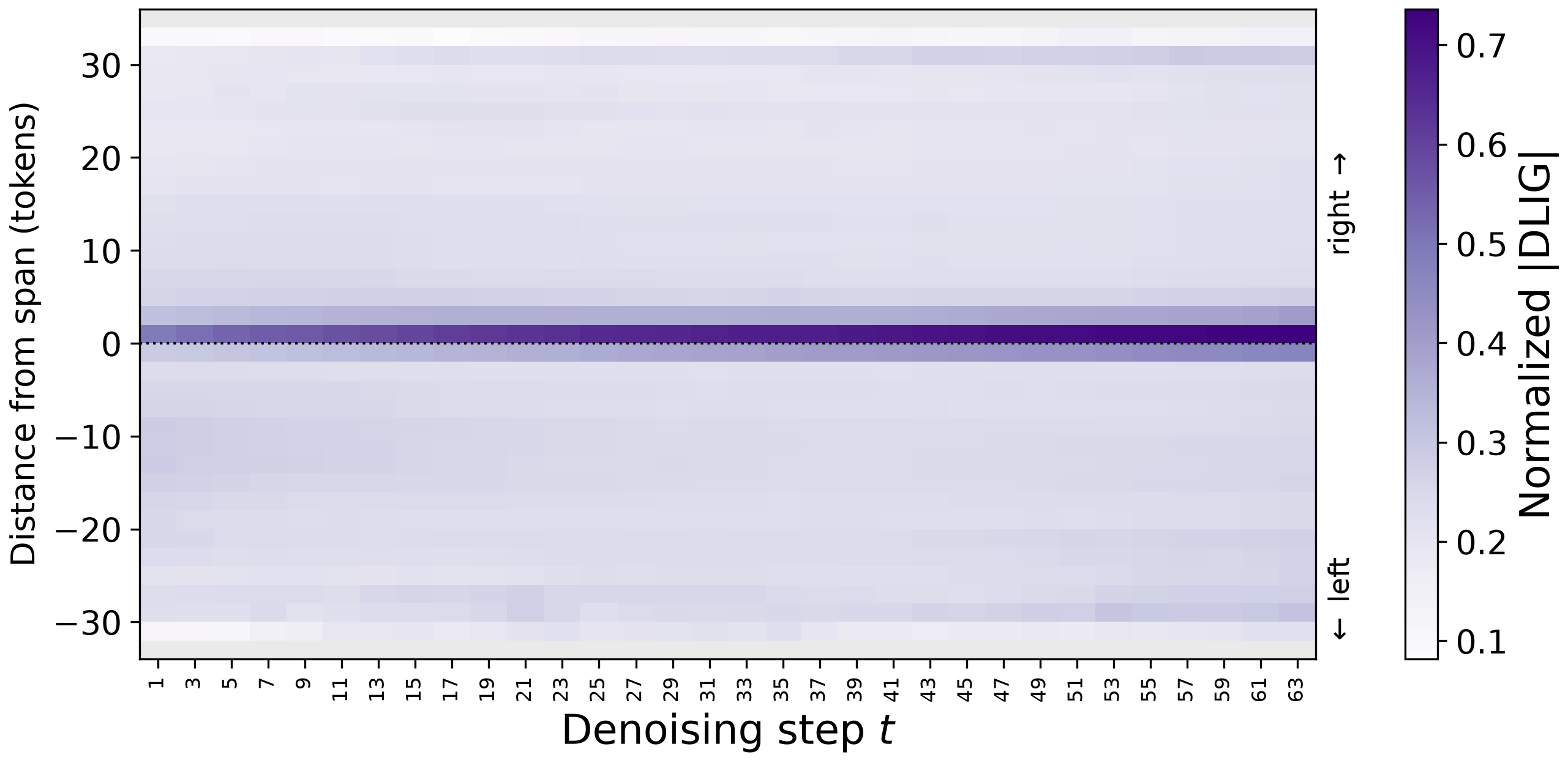}
\caption{Fixed-target robustness: mean normalized DLIG magnitude by denoising step and signed distance from the span (fixed gold target). The bidirectional, left-skewed shape matches the self-generated profile (Fig.~\ref{fig:roc-profile}); only the overall magnitude differs.}
\label{fig:roc-profile-fixed}
\end{figure}

\paragraph{The descriptive attribution profile is framing-robust.} The bidirectional, left-skewed profile of \S~\ref{sec:rocstories} is unchanged under the fixed target (Fig.~\ref{fig:roc-profile-fixed}): mass peaks on both sides, the left context carries higher magnitude, and magnitude decays across the trajectory. This shape holds regardless of infill quality, consistent with the quality-neutrality of context mass (Appendix~\ref{app:roc-mass-auc}). The overall magnitudes differ between framings, but the shape of the attribution field does not.

\paragraph{Directional balance under the fixed target.} The right/left context-mass ratio is again near-constant across quality, so quality tracks directional balance under neither framing. The framings differ only in where the balance sits: the fixed gold target pulls attribution mildly toward the right context, which constrains the gold continuation, while the self-generated target is mildly left-heavy (Table~\ref{tab:roc-gold-fixed-target-stats} and Table~\ref{tab:roc-mass-auc-summary} give the matched ratios).

\section{Statistical Evaluation}
\label{app:stats}

This section presents the full statistical tests for every experiment in \S~\ref{sec:wic}, \S~\ref{sec:prosqa} and \S~\ref{sec:rocstories}. All quantities are computed per example and, where noted, pooled over the analyzed grid of denoising steps. We test each quantity against the null appropriate to its form: a \emph{one-sample} claim (a mass or gap is nonzero) is tested with the Wilcoxon signed-rank test against zero; a \emph{two-group} claim (a quantity discriminates success from fail, or predicted-Yes from predicted-No) is reported as the Mann--Whitney AUC, $U/(n_1 n_2)$, where $0.5$ is chance and values below $0.5$ mean the quantity is \emph{larger} on the second group, with the two-sided $p$; and a \emph{continuous-association} claim (a quantity tracks a graded outcome) is reported as a Pearson correlation, with $95\%$ bootstrap confidence intervals over the resampling unit. Where a genuine effect must be separated from problem-difficulty confounds, we additionally report Spearman correlations against difficulty covariates, group statistics recomputed within difficulty strata (sample-size-weighted average), and standardized logistic or linear regressions with difficulty covariates as controls (Wald $p$ per coefficient); structure-sensitive masses are compared against a structure-matched random-edge null. Significance markers follow the convention $^{*}p{<}0.05$, $^{**}p{<}0.01$, $^{***}p{<}0.001$.

\begin{table}[t]
\centering
\small
\caption{Full statistical results for word-sense discrimination on WiC (\S~\ref{sec:wic}).}
\label{tab:wic-stats}
\begin{tabularx}{\linewidth}{
    @{}
    >{\raggedright\arraybackslash}p{0.30\linewidth}
    >{\raggedright\arraybackslash}X
    >{\raggedleft\arraybackslash}p{0.20\linewidth}
    @{}
}
\toprule
\textit{Analysis} & \textit{Statistic} & \textit{Value} \\
\midrule
\multirow[t]{3}{=}{Overall task accuracy}
& Overall accuracy ($n{=}500$)
& $63.6\%$ \\
& Same-sense (Yes) accuracy ($n{=}250$)
& $90.8\%$ \\
& Different-sense (No) accuracy ($n{=}250$)
& $36.4\%$ \\
\midrule
\multirow[t]{4}{=}{Behavioral divergence: predicted-Yes vs.\ predicted-No}
& Attribution depth (centroid): mean pred-Yes / pred-No
& $7.017$ / $7.014$ \\
& Attribution depth: AUC / $p$
& $0.503$ / $0.916$\ \ (n.s.) \\
& Commitment timing: mean pred-Yes / pred-No
& $0.073$ / $0.271$ \\
& Commitment timing: AUC / $p$
& $0.342$ / $<10^{-4}\;^{***}$ \\
\midrule
\multirow[t]{4}{=}{Hidden structure within predicted-Yes (unsupervised)}
& Attribution depth: BIC 1-comp / 2-comp ($n{=}394$)
& $499.4$ / $512.1$ \\
& Attribution depth: modes ($w$)
& $6.914\,(0.66)$ / $7.215\,(0.34)$ \\
& Commitment timing: BIC 1-comp / 2-comp ($n{=}394$)
& $-75.2$ / $-57.3$ \\
& Commitment timing: modes ($w$)
& $0.073\,(0.50)$ / $0.073\,(0.50)$ \\
\bottomrule
\end{tabularx}
\end{table}

\begin{table}[t]
\centering
\small
\caption{Fraction of WiC test examples with zero prompt-token attribution mass ($M_t \le 10^{-9}$) at each denoising step (DiffuGPT-M, $n{=}499$ parseable predictions). Zero-mass steps occur before the Yes/No span commits, when the self-generated target carries no attributable signal, and are excluded from the attribution-depth centroid. Aggregate over all 32 steps: $22.3\%$.}
\label{tab:wic-stepwise-zero-mass}
\begin{tabular}{@{}rr|rr@{}}
\toprule
$t$ & zero-mass \% & $t$ & zero-mass \% \\
\midrule
1  & 93.0 & 33 & 7.8 \\
3  & 84.2 & 35 & 7.4 \\
5  & 75.6 & 37 & 6.2 \\
7  & 67.1 & 39 & 5.4 \\
9  & 59.7 & 41 & 4.4 \\
11 & 51.1 & 43 & 3.6 \\
13 & 44.3 & 45 & 2.6 \\
15 & 37.5 & 47 & 2.0 \\
17 & 31.3 & 49 & 1.4 \\
19 & 26.1 & 51 & 1.2 \\
21 & 22.6 & 53 & 0.8 \\
23 & 20.2 & 55 & 0.6 \\
25 & 17.8 & 57 & 0.4 \\
27 & 15.4 & 59 & 0.0 \\
29 & 12.6 & 61 & 0.0 \\
31 & 9.6  & 63 & 0.0 \\
\bottomrule
\end{tabular}
\end{table}

\begin{table}[t]
\centering
\small
\caption{Full statistical results for multi-hop graph reasoning (\S~\ref{sec:prosqa}).}
\label{tab:prosqa-stats}
\begin{tabularx}{\linewidth}{
    @{}
    >{\raggedright\arraybackslash}p{0.30\linewidth}
    >{\raggedright\arraybackslash}X
    >{\raggedleft\arraybackslash}p{0.20\linewidth}
    @{}
}
\toprule
\textit{Analysis} & \textit{Statistic} & \textit{Value} \\
\midrule
\multirow[t]{4}{=}{Attribution concentrates on the gold path (successes, pooled)}
& Precision $-$ chance
& $+0.150$ \\
& Wilcoxon $p$
& $<10^{-15}\;^{***}$ \\
& Random-edge null: mean
& $-0.020$ \\
& Random-edge null: 95th percentile
& $+0.075$ \\
\midrule
\multirow[t]{6}{=}{Faithfulness on fails: mass tracks the model's cited chain}
& Cited chain vs.\ other facts: per-token gap ($n{=}66$)
& $+4.77$ \\
& Cited chain vs.\ other facts: Wilcoxon $p$
& $3.5{\times}10^{-6}\;^{***}$ \\
& Gold vs.\ other facts (success mirror): per-token gap ($n{=}286$)
& $+3.50$ \\
& Gold vs.\ other facts (success mirror): Wilcoxon $p$
& $<10^{-15}\;^{***}$ \\
& Divergent-token subset: chain vs.\ gold ($n{=}42$)
& $-1.46$ \\
& Divergent-token subset: paired Wilcoxon $p$
& $0.52$\ \ (n.s.) \\
\midrule
\multirow[t]{12}{=}{Contrastive magnitude anti-predicts correctness}
& Mean per-token $|d|$: success / fail
& $3.55$ / $4.13$ \\
& Mean per-token $|d|$: AUC / $p$
& $0.440$ / $0.124$\ \ (n.s.) \\
& Difficulty checks: Spearman$(|d|,k)$
& $+0.06$\ \ (n.s.) \\
& Difficulty checks: Spearman$(|d|,\text{size})$
& $-0.33$ \\
& Within-$k$ AUC: $k{=}2$ ($n{=}120{+}27$)
& $0.365\;^{*}$ \\
& Within-$k$ AUC: $k{=}3$ ($n{=}116{+}33$)
& $0.472$\ \ (n.s.) \\
& Within-$k$ AUC: $k{=}4$ ($n{=}45{+}8$)
& $0.511$\ \ (n.s.) \\
& Within-$k$ AUC: weighted average
& $0.433$ \\
& Logistic ($+\,k$, size): $\beta_{|d|}$ / $p$
& $-0.203$ / $0.125$\ \ (n.s.) \\
& Logistic ($+\,k$, size): $\beta_k$ / $p$
& $+0.033$ / $0.81$\ \ (n.s.) \\
& Logistic ($+\,k$, size): $\beta_{\text{size}}$ / $p$
& $+0.044$ / $0.76$\ \ (n.s.) \\
& Per-step AUC: $t{=}1,7,13,19,25,31,37,43,49,54,60$
& $.435,.424^{*},.430,$\newline $.445,.439,.437,$\newline $.432,.416^{*},.421^{*},$\newline $.419^{*},.419^{*}$ \\
\midrule
\multirow[t]{4}{=}{Attribution location is group-invariant}
& Precision $-$ chance: success / fail
& $+0.150$ / $+0.138$ \\
& AUC / $p$
& $0.524$ / $0.53$\ \ (n.s.) \\
& Joint logistic: $\beta_{\text{prec}}$ / $p$
& $+0.064$ / $0.64$\ \ (n.s.) \\
& Joint logistic: $\beta_{|d|}$ / $p$
& $-0.213$ / $0.089$\ \ (n.s.) \\
\bottomrule
\end{tabularx}
\end{table}

\begin{table}[t]
\centering
\small
\caption{Full statistical results for the ROCStories infilling attribution analysis (\S~\ref{sec:rocstories}), self-generated target mode.}
\label{tab:infill-stats}
\begin{tabularx}{\linewidth}{
    @{}
    >{\raggedright\arraybackslash}p{0.30\linewidth}
    >{\raggedright\arraybackslash}X
    >{\raggedleft\arraybackslash}p{0.20\linewidth}
    @{}
}
\toprule
\textit{Analysis} & \textit{Statistic} & \textit{Value} \\
\midrule
\multirow[t]{2}{=}{ROUGE-1 distribution (self-generated span vs.\ gold)}
& Mean / median
& $0.212$ / $0.200$ \\
& Range
& $[0.000,\ 0.833]$ \\
\midrule
\multirow[t]{6}{=}{Self-target attribution tracks output quality when pooled}
& Ratio(right/left) vs.\ ROUGE: Pearson $r$ ($n{=}28514$)
& $0.027$ \\
& Mean attribution ratio (right/left)
& $0.984$ \\
& Total-mass AUC (pooled, high vs.\ low ROUGE)
& $0.510\;[0.503,0.517]$ \\
& Total-mass $p$
& $3.4{\times}10^{-3}\;^{**}$ \\
& Per-step AUC: $t{=}1,17,33,49,63$
& $0.501,\,0.517,\,0.533,$\newline $0.507,\,0.486$ \\
& Per-step AUC range across all 32 steps
& $[0.485,\ 0.533]$ \\
\midrule
\multirow[t]{3}{=}{Attribution is left-skewed (paired, pooled over stories $\times$ steps)}
& Mean left mass / mean right mass ($n{=}28514$ story-step cells)
& $34.18$ / $32.47$ \\
& Paired Wilcoxon $p$ (left vs.\ right)
& $1.9{\times}10^{-204}\;^{***}$ \\
& Fraction of cells left $>$ right / left $<$ right
& $57.8\%$ / $42.2\%$ \\
\midrule
Reliance strengthens late in denoising
& Mean total mass, $t{=}39$ vs.\ $t{=}63$
& $59.36$ / $63.85$ \\
\bottomrule
\end{tabularx}
\end{table}

\begin{table}[t]
\centering
\small
\caption{Mean total context attribution mass per denoising step, ROCStories infilling (self-generated target).}
\label{tab:infill-stepwise-mass}
\begin{tabular}{@{}rrr|rrr@{}}
\toprule
$t$ & $n$ & mean mass & $t$ & $n$ & mean mass \\
\midrule
1  & 124 & 101.96 & 33 & 995  & 60.93 \\
3  & 361 & 99.76  & 35 & 995  & 60.11 \\
5  & 506 & 96.04  & 37 & 998  & 59.67 \\
7  & 622 & 91.08  & 39 & 1000 & 59.36 \\
9  & 719 & 87.21  & 41 & 1000 & 59.00 \\
11 & 792 & 83.71  & 43 & 1000 & 58.96 \\
13 & 845 & 81.09  & 45 & 1000 & 58.86 \\
15 & 880 & 76.95  & 47 & 1000 & 59.00 \\
17 & 914 & 74.68  & 49 & 1000 & 59.06 \\
19 & 932 & 72.48  & 51 & 1000 & 59.24 \\
21 & 949 & 69.72  & 53 & 1000 & 59.80 \\
23 & 963 & 67.89  & 55 & 1000 & 60.66 \\
25 & 970 & 65.92  & 57 & 1000 & 61.40 \\
27 & 976 & 64.37  & 59 & 1000 & 61.95 \\
29 & 983 & 62.87  & 61 & 1000 & 62.93 \\
31 & 990 & 61.85  & 63 & 1000 & 63.85 \\
\bottomrule
\end{tabular}
\end{table}

Table~\ref{tab:wic-stats} reports full statistical results for the WiC experiment (\S~\ref{sec:wic}); Table~\ref{tab:wic-stepwise-zero-mass} gives the per-step zero-mass fractions referenced from the attribution-depth footnote in \S~\ref{sec:wic}. Table~\ref{tab:prosqa-stats} reports full statistical results for the ProsQA experiment (\S~\ref{sec:prosqa}). Tables~\ref{tab:infill-stats} and~\ref{tab:infill-stepwise-mass} report full statistical results for the ROCStories infilling experiment (\S~\ref{sec:rocstories}).

\section{Model, Datasets and Training Details}
\label{app:model_training}

All three tasks finetune DiffuGPT-M~\cite{gong2025scaling}, a masked-diffusion adaptation of GPT-2-medium~\cite{radford2019language} ($\sim$355M params). Table~\ref{tab:wic-training} gives the data and fine-tuning configuration for WiC; Table~\ref{tab:prosqa-training} gives them for ProsQA. Task-level performance for both is reported in Table~\ref{tab:task-performance}; generation and evaluation settings for all three tasks are given in Table~\ref{tab:gen-eval-config}.

\paragraph{Attribution step grids.} WiC and ROCStories infilling attribute at all 32 odd steps $t\in\{1,3,5,\ldots,63\}$ of the $T{=}64$ trajectory. ProsQA attributes at 11 steps $t\in\{1,7,13,19,25,31,37,43,49,54,60\}$: each ProsQA cell scores a full per-token contrastive attribution over the prompt's fact graph, markedly more expensive than the single-target scoring used for WiC/infill, so the grid is subsampled for tractability.

\paragraph{WiC.} WiC~\cite{pilehvar2019wic} ships a 5,428-train / 638-val split; the official test labels are not public. We therefore carve a label-stratified test set from train, yielding the 4,928 / 638 / 500 split of Table~\ref{tab:wic-training}. Each instance is cast in a wrapper querying whether the target word shares its sense across the two sentences (prompt format in Table~\ref{tab:wic-training}), with a binary \texttt{Yes}/\texttt{No} target. Because the answer tokens share no vocabulary with the disambiguating context, the task cannot be solved by surface-form overlap.

\paragraph{ProsQA.} ProsQA~\citep{hao2024training} is a synthetic multi-hop graph-reasoning benchmark. Each question states an ontology as a set of typing facts—including distractors unrelated to the gold reasoning path—assigns the queried subject to a root concept, and poses a binary single-correct choice (``Is $s$ a $A$ or $B$?''), with the gold option reachable from the root by a $k$-hop chain of stated facts. We use the 17,886 / 300 / 500 split (test hop mix in Table~\ref{tab:prosqa-training}).

\paragraph{ROCStories infilling.} \citet{gong2025scaling} evaluate DiffuGPT-M on ROCStories~\cite{mostafazadeh2016corpus} infilling; we adopt the same setup, generating sentence 3 from sentences 1--2 (left context) and 4--5 (right context) over 1000 evaluation samples. The task is a natural DLIG target because the generated span attends bidirectionally, letting DLIG resolve how attribution mass distributes between left and right context and how that distribution evolves over denoising steps.

\begin{table}[t]
\centering
\small
\caption{Data and fine-tuning configuration for the WiC word-sense disambiguation task.}
\label{tab:wic-training}
\begin{tabularx}{\linewidth}{
    @{}
    >{\raggedright\arraybackslash}p{0.22\linewidth}
    >{\raggedright\arraybackslash}p{0.28\linewidth}
    >{\raggedright\arraybackslash}X
    @{}
}
\toprule
\textit{Section} & \textit{Parameter} & \textit{Value} \\
\midrule
\multirow[t]{4}{=}{Data (WiC)}
& Train / val / test & 4{,}928 / 638 / 500 \\
& Test split & label-stratified hold-out from train (official test labels not public) \\
& Prompt format & ``Sentence 1: \ldots\ Sentence 2: \ldots\ Does the word ``\textless w\textgreater'' have the same meaning in both sentences?'', empty system prompt \\
& Target & binary \texttt{\#\#\# Yes.} / \texttt{\#\#\# No.} (\texttt{<eos>}-terminated) \\
\midrule
\multirow[t]{8}{=}{Fine-tuning}
& Objective & diffusion SFT (\texttt{ddm-sft}), full fine-tuning \\
& Answer-token loss weight & $10\times$ on the \texttt{Yes}/\texttt{No} token (counteracts \texttt{Yes}-bias) \\
& Attention / readout & bidirectional retained (\texttt{anneal\_steps=1}); shifted (\texttt{shift=true}) \\
& Optimizer / LR & AdamW, $1{\times}10^{-5}$ \\
& Schedule / warmup & cosine, $0.03$ \\
& Epochs & 40 \\
& Batch size & 8 per device $\times$ 4 grad.\ accum.\ steps \\
& Max sequence length & 128 (\texttt{cutoff\_len}) \\
\bottomrule
\end{tabularx}
\end{table}

\begin{table}[t]
\centering
\small
\caption{Data and fine-tuning configuration for the ProsQA multi-hop task.}
\label{tab:prosqa-training}
\begin{tabularx}{\linewidth}{
    @{}
    >{\raggedright\arraybackslash}p{0.22\linewidth}
    >{\raggedright\arraybackslash}p{0.28\linewidth}
    >{\raggedright\arraybackslash}X
    @{}
}
\toprule
\textit{Section} & \textit{Parameter} & \textit{Value} \\
\midrule
\multirow[t]{3}{=}{Data (ProsQA)}
& Train / val / test & 17{,}886 / 300 / 500 \\
& Prompt format & raw ``Is \textless subj\textgreater\ a $A$ or $B$?'', empty system prompt \\
& Test hop mix ($k$) & 2:202,\ 3:217,\ 4:69,\ 5:12 \\
\midrule
\multirow[t]{7}{=}{Fine-tuning}
& Objective & diffusion SFT (\texttt{ddm-sft}), full fine-tuning \\
& Attention / readout & bidirectional retained (\texttt{anneal\_steps=1}); shifted (\texttt{shift=true}) \\
& Optimizer / LR & AdamW, $1{\times}10^{-5}$ \\
& Schedule / warmup & cosine, $0.03$ \\
& Epochs & 10 \\
& Batch size & 8 per device $\times$ 4 grad.\ accum.\ steps \\
& Max sequence length & 640 (\texttt{cutoff\_len}) \\
\bottomrule
\end{tabularx}
\end{table}

\begin{table}[t]
\centering
\small
\caption{Generation and evaluation configuration, shared sampler settings and per-task overrides.}
\label{tab:gen-eval-config}
\begin{tabularx}{\linewidth}{
    @{}
    >{\raggedright\arraybackslash}p{0.30\linewidth}
    >{\raggedright\arraybackslash}X
    @{}
}
\toprule
\textit{Parameter} & \textit{Value} \\
\midrule
Denoising steps $T$ & 64 \\
Reveal schedule & random-reveal, $p_{\to x_0}{=}1/(t{+}1)$ \\
Sampling & nucleus, $p{=}0.9$; temperature $1.0$ \\
Seed & 42 \\
Precision & fp32 \\
\midrule
Max new tokens (WiC) & 8 \\
Max new tokens (ProsQA) & 64 \\
Max new tokens (ROCStories infill) & oracle span length (gold sentence's own token length) \\
\bottomrule
\end{tabularx}
\end{table}
\end{document}